\documentclass[11pt]{article}

\usepackage[final]{acl}

\usepackage{times}
\usepackage{latexsym}

\usepackage[T1]{fontenc}

\usepackage[utf8]{inputenc}

\usepackage{microtype}

\usepackage{inconsolata}

\usepackage{graphicx}

\usepackage{amsmath}
\usepackage{paralist}
\usepackage{booktabs}
\usepackage{algorithm}
\usepackage{algpseudocode}

\usepackage{xcolor}
\definecolor{structcolor}{RGB}{214,220,243}
\definecolor{contentcolor}{RGB}{254,227,226}
\definecolor{answercolor}{RGB}{240,218,15}

\usepackage[many]{tcolorbox}
\usepackage{lipsum}
\usepackage{booktabs}
\usepackage{newunicodechar}
\usepackage{tabularx}
\newunicodechar{√}{$\surd$} 
\newtcolorbox{onecolprompt}[1]{
    colback=gray!5!white,
    colframe=gray!70!black,
    fonttitle=\bfseries,
    title=#1,
    breakable,
    enhanced,
    before upper={\ttfamily\small\obeyspaces},
    width=\columnwidth,
    left=2pt, right=2pt
}
\newtcolorbox{twocolprompt}[1]{
    colback=gray!5!white,
    colframe=gray!70!black,
    fonttitle=\bfseries,
    title=#1,
    breakable,
    enhanced,
    before upper={\ttfamily\small\obeyspaces},
    width=\textwidth,
    enlarge left by=-\columnsep/2,
    enlarge right by=-\columnsep/2,
    left=2pt, right=2pt
}

\title{Equipping Retrieval-Augmented Large Language Models with Document Structure Awareness}

\author{
 \textbf{Lingnan Xu\textsuperscript{$\dagger$,$\ddagger$}},
 \textbf{Chong Feng\textsuperscript{$\dagger$,$\S$}\thanks{Corresponding Author}},
 \textbf{Kaiyuan Zhang\textsuperscript{$\dagger$}},
\\
 \textbf{Liu Zhengyong\textsuperscript{$\ddagger$}},
 \textbf{Wenqiang Xu\textsuperscript{$\ddagger$}},
 \textbf{Fanqing Meng\textsuperscript{$\dagger$}}
\\
 \textsuperscript{$\dagger$}School of Computer Science, Beijing Institute of Technology \textsuperscript{$\ddagger$}Ant Group
\\
 \textsuperscript{\S}Southeast Academy of Information Technology, Beijing Institute of Technology
\\
 \small{
  \{xln, fengchong, zky, mengfanqing\}@bit.edu.cn,
 }
\\
 \small{
  \{xulingnan.xln, liuzhengyong.lzy, yugong.xwq\}@antgroup.com
 }
}

\begin{document}
\maketitle
\begin{abstract}
While large language models (LLMs) demonstrate impressive capabilities, their reliance on parametric knowledge often leads to factual inaccuracies. Retrieval-Augmented Generation (RAG) mitigates this by leveraging external documents, yet existing approaches treat retrieved passages as isolated chunks, ignoring valuable structure that is crucial for document organization. Motivated by this gap, we propose \textit{\textbf{R}etrieve-\textbf{D}ocument\textbf{R}oute-\textbf{R}ead} (\textbf{RDR\textsuperscript{2}}), a novel framework that explicitly incorporates structural information throughout the RAG process. RDR\textsuperscript{2} employs an LLM-based router to dynamically navigate document structure trees, jointly evaluating content relevance and hierarchical relationships to assemble optimal evidence. Our key innovation lies in formulating document routing as a trainable task, with automatic action curation and structure-aware passage selection inspired by human reading strategies. Through comprehensive evaluation on five challenging datasets, RDR\textsuperscript{2} achieves state-of-the-art performance, demonstrating that explicit structural awareness significantly enhances RAG systems' ability to acquire and utilize knowledge, particularly in complex scenarios requiring multi-document synthesis.\footnote{Code \& data: \url{https://github.com/XuLingnan/RDR2}}
\end{abstract}

\section{Introduction}

Large language models (LLMs) \citep{brown2020language} have demonstrated remarkable capabilities across a wide range of natural language processing (NLP) tasks, yet even state-of-the-art models continue to generate factually incorrect responses \citep{mallen-etal-2023-trust, min-etal-2023-factscore, ji2023survey} despite their growing scale and capability \citep{ouyang2022training}. Retrieval-Augmented Generation (RAG) \citep{lewis2020retrieval, guu2020retrieval, borgeaud2022improving} addresses these limitations through a \textit{Retrieve-and-Read} paradigm, which first retrieves relevant passages then uses them as context for generation \citep{lewis2020retrieval, izacard-grave-2021-leveraging, jiang-etal-2022-retrieval, shi-etal-2024-replug}. This approach combines the strengths of information retrieval and generative models, proving particularly effective for atomic-fact question answering (QA) \citep{joshi-etal-2017-triviaqa, thorne-etal-2018-fever, kwiatkowski-etal-2019-natural, mallen-etal-2023-trust} where \textit{a single precise retrieval suffices to answer clear information needs}.

Recent advances in RAG have extended its capabilities to complex knowledge-intensive scenarios requiring multi-perspective responses, particularly for factual-inductive queries that demand coherent synthesis of multiple knowledge fragments \citep{fan-etal-2019-eli5, stelmakh-etal-2022-asqa, amouyal-etal-2023-qampari}. However, current RAG frameworks process retrieved passages as isolated chunks, discarding their inherent document structure - a limitation stemming from both \textit{structure-agnostic pipeline design} and the \textit{flat-context paradigm of standard retrieval methods}. 

While fixed chunking ensures retrieval efficiency, it restricts query-adaptive content selection, discarding the document's native organization which humans naturally exploit for information navigation and relational reasoning. At the reading phase, retrieved passages are simply ordered by relevance scores, potentially disrupting their original sequence in the source document. Even with useful information, this loss of structural priors forces the model to implicitly reconstruct relationships that were explicitly encoded in the source hierarchy. This structural blindness constrains RAG's knowledge acquisition and synthesis capabilities.

In this paper we ask: \textit{can LLMs leverage document structural information}, and \textit{can RAG systems benefit from such structural awareness}? We propose \textit{\textbf{R}etrieve-\textbf{D}ocument\textbf{R}oute-\textbf{R}ead} (\textbf{RDR\textsuperscript{2}}), where a structure-aware LLM performs document routing through three actions inspired by how humans selectively read sections, expand promising headings, and skip irrelevant parts when browsing articles. Through this process, RDR\textsuperscript{2} dynamically assembles query-oriented passages for better knowledge acquisition and utilization.

\begin{figure}[h]
  \includegraphics[width=\columnwidth]{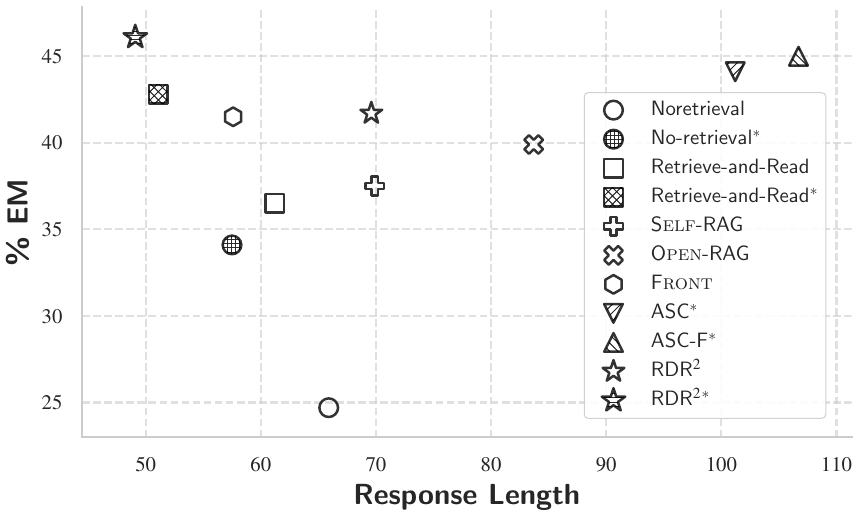}
  \caption{Performance comparison on ASQA, where RDR\textsuperscript{2} achieves the highest Exact Match (EM) score while generating the most concise responses. Readers are based on either Llama-2-13B or ChatGPT (*).}
  \label{fig:intro}
\end{figure}

We evaluate RDR\textsuperscript{2} on five QA datasets representing diverse formats: TriviaQA \citep{joshi-etal-2017-triviaqa} (single-answer), HotpotQA \citep{yang-etal-2018-hotpotqa} (multi-hop), QAMPARI \citep{amouyal-etal-2023-qampari} (list-style), ASQA \citep{stelmakh-etal-2022-asqa} (ambiguous), and ELI5 \citep{fan-etal-2019-eli5} (in-depth). As shown in Figure~\ref{fig:intro}, RDR\textsuperscript{2} achieves new state-of-the-art results with only the router trained on questions from the ASQA training set (without answer supervision), while keeping the retriever and reader off-the-shelf. Additionally, RDR\textsuperscript{2} enables test-time scaling without weight updates and demonstrates generalization across different RAG components (i.e., retrievers and readers).

Our main contributions are:
\begin{compactitem}
\item The proposal of RDR\textsuperscript{2}, the first RAG framework explicitly incorporates document structure throughout the retrieval and reading process, to enhance both knowledge acquisition and utilization;
\item A novel formulation of document routing as a trainable task, with an automatic action curation pipeline and LLM-based router training;
\item Comprehensive experiments on five datasets establishing RDR\textsuperscript{2}'s consistent superiority over state-of-the-art methods.
\end{compactitem}

\section{Related Work}

Retrieval-Augmented Generation \citep{lewis2020retrieval, guu2020retrieval, borgeaud2022improving} (RAG) augments language models with non-parametric knowledge through retrieved passages, demonstrating significant improvements in knowledge-intensive tasks \citep{ram-etal-2023-context, asai-etal-2023-retrieval}. The standard \textit{Retrieve-and-Read} framework operates in two stages: (1) a dense retriever, typically employing a bi-encoder architecture \citep{karpukhin-etal-2020-dense, ni-etal-2022-large, wang-etal-2024-improving-text}, retrieves passages relevant to the input question, and (2) an LM reader processes these passages, either as an off-the-shelf model \citep{zhou2024language, li2025search} or through task-specific fine-tuning \cite{izacard2023atlas, lin2023ra, jain-etal-2023-1, luo2024reasoning, gan2024similarity}, to generate grounded responses. While effective for simple tasks with clear information needs, RAG systems show limitations in complex scenarios, necessitating more advanced methods.

\textbf{Knowledge Acquisition.} To achieve more comprehensive knowledge acquisition, recent works develop enhanced retrieval mechanisms. 
FLARE \citep{jiang-etal-2023-active} prompts an LLM to actively decide when and what to retrieve based on the model's confidence (i.e., token probabilities). 
\citet{ma-etal-2023-query} introduces query rewriting to bridge the gap between user questions and retrieval requirements. 
CoRAG \citep{wang2025chain} fine-tunes an LLM to generate intermediate retrieval chains, enabling step-by-step multi-hop querying. 
Unlike prior works that focus on pre-retrieval query optimization, our approach enhances knowledge acquisition through post-retrieval document routing - iteratively exploring document hierarchies to uncover useful information.

\textbf{Knowledge Utilization.} For knowledge utilization, effective RAG requires critical evaluation and integration of retrieved knowledge. 
S{\small ELF}-RAG \citep{asai2023self} fine-tunes LLMs to critique retrieved passages via self-reflection, assessing their relevance, supportiveness, and utility. 
RankRAG \citep{yu2024rankrag} instruction-tunes a single LLM for the dual purpose of context ranking and answer generation, improving end-to-end knowledge grounding. 
Departing from static chunk filtering, our method dynamically assembles node-level information units within document hierarchy, achieving both structural integrity and adaptive flexibility.

\begin{figure*}[t]
  \includegraphics[width=\linewidth]{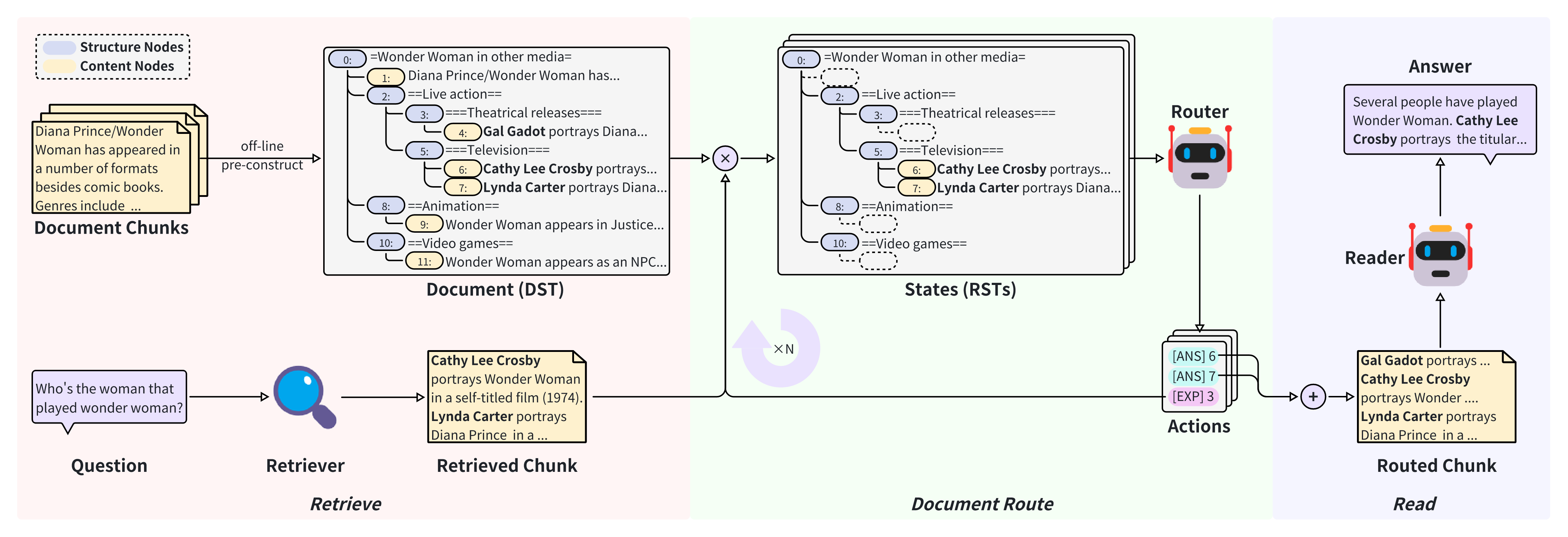}
  \caption {Overwiew of the RDR\textsuperscript{2} framework. RDR\textsuperscript{2} extends standard \textit{Retrieve-and-Read} with document-structure-aware routing for iterative, fine-grained knowledge retrieval. \textit{Retrieve}: input question $q$, output retrieved chunks $C_{re}$; \textit{Document Route}: input $q$, $C_{re}$ and corresponding documents $D$, output routed chunks $C_{ro}$; \textit{Read}: input $q$ and $C_{ro}$, output final answer $a$.}
  \label{fig:rdr2_framework}
\end{figure*}

\textbf{Structural Information.} Several approaches have attempted to incorporate structural information into RAG frameworks. 
GraphRAG \citep{edge2024local} processes documents into a knowledge graph with hierarchical community summaries, establishing a RAG paradigm distinct from semantic retrieval over flat text chunk. 
RAPTOR \citep{sarthi2024raptor} constructs hierarchical document embeddings through recursive node-level clustering and summarization, capturing progressively abstracted semantic content across tree levels. 
While existing approaches \textit{offline-encode} hierarchical information into \textit{fixed representations} (e.g., summaries or embeddings), our framework \textit{online-perceives} document structure through \textit{dynamic routing}.

\section{Methodology}

In this section, we present RDR\textsuperscript{2} (\textit{Retrieve-DocumentRoute-Read}), a novel framework that endows the retrieval-augmented systems with explicit awareness of document structure. We first introduce the overview of our framework. Then we define tree structures to represent the document hierarchy, ensuring stable scope and adaptive contextual focus. Finally, we introduce the document routing task and the scheme  for training a structure-aware LLM router.

\subsection{\textit{Retrieve-DocumentRoute-Read}}
\label{sec:rdr2_framework}

As illustrated in Figure~\ref{fig:rdr2_framework}, the RDR\textsuperscript{2} framework consists of three stages: 

\textbf{Retrieve.} Given an input question $q$ and a datastore $\mathcal{D}$, the $\mathrm{Retriever}$ retrieves the top-$k$ most relevant chunks $C_{re} = \{c_{re}^{(1)}, \cdots, c_{re}^{(k)}\}$, along with their originating documents $D = \{d_1, \cdots, d_k\}$.
\begin{equation}
  \label{eq:retriever}
  {\{\langle c_{re}^{(i)}, d_i \rangle\}}_{i=1}^{k} = \mathrm{Retriever}(q, \mathcal{D})
\end{equation}

\begin{figure*}[t]
  \includegraphics[width=\linewidth]{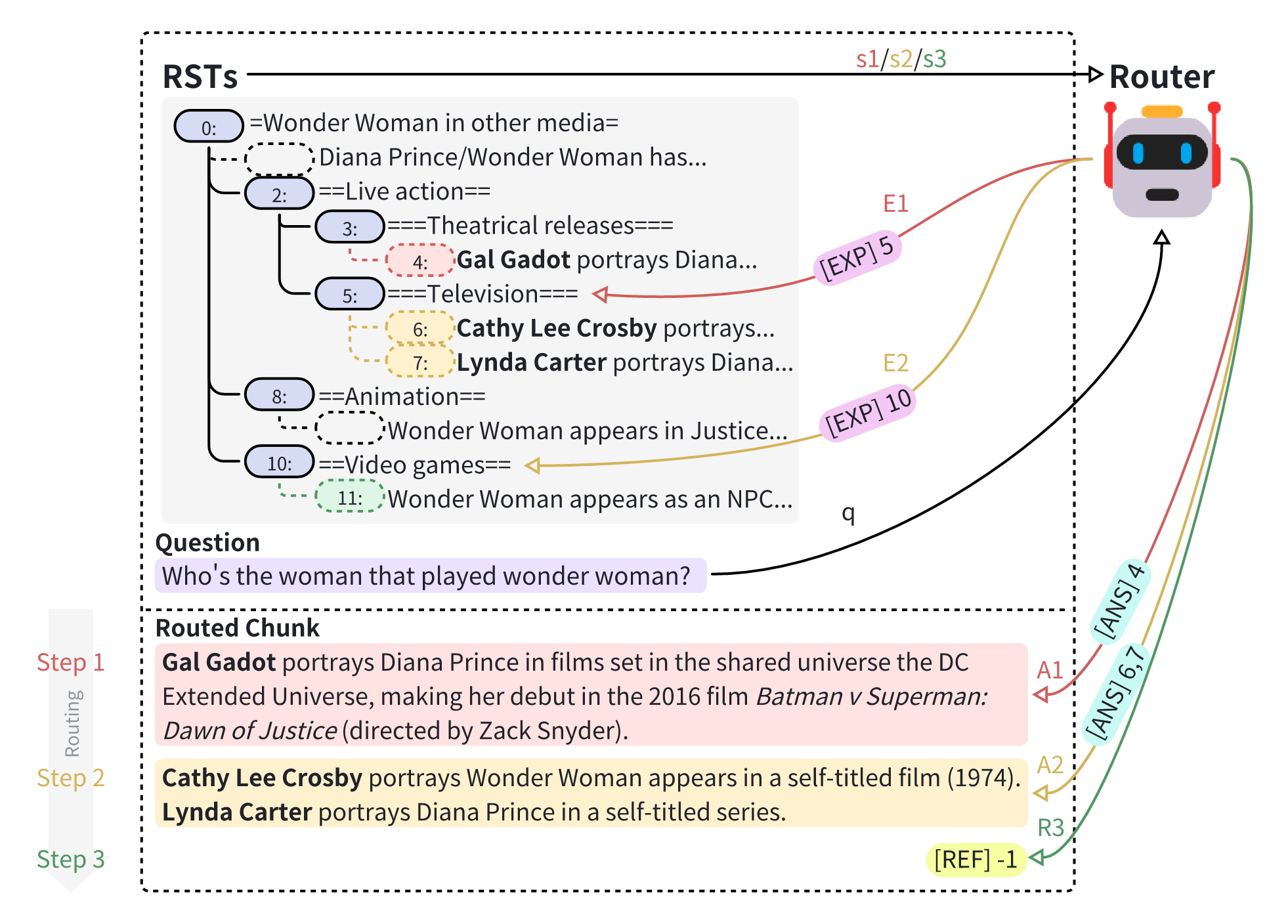}
  \caption {Workflow of the routing module. Given a user input $q$ and a document structure tree (Section~\ref{sec:dst}) anchored by retrieved chunks, RDR\textsuperscript{2} maintains a retrieval subtree $s$ where: (i) all structure nodes persist, (ii) only content nodes under currently selected headings are expanded (previous fold). At step $t$, the router generates action $\{\langle a_{j}^{(t)}, p_{j}^{(t)} \rangle\}_{j=1}^{n_t} = \mathrm{Router}(q, s_t)$ to: (a) select useful content nodes, (b) unfold a promising structure node, or (c) stops routing.
  }
  \label{fig:routing_module}
\end{figure*}

\textbf{Document Route.} This stage transforms chunk-wise retrieved results into document-wise routed chunks $C_{ro} = \{c_{ro}^{(1)}, \cdots, c_{ro}^{(m)}\}$ through an iterative process, where an LLM-based $\mathrm{Router}$ selectively expand relevant sections while maintaining awareness of the document's organizational framework. At each step $t$, the $\mathrm{Router}$ takes the question $q$ and the current routing state $s_i^{(t)}$ to decide a series of actions, where each element consists of a ternary action tag $a_{ij}^{(t)} \in \{\mathsf{[ANS]}, \mathsf{[EXP]}, \mathsf{[REF]}\}$\footnote{$\mathsf{[ANS]}$: extracting useful contents to \textit{answer}; $\mathsf{[EXP]}$: unfolding promising titles to \textit{expand}; $\mathsf{[REF]}$: stopping the routing process to \textit{refuse}.}, along with a selected passage node $p_{ij}^{(t)}$.
\begin{equation}
  \{\langle a_{ij}^{(t)}, p_{ij}^{(t)} \rangle\}_{j=1}^{n_i^{(t)}} = \mathrm{Router}(q, s_i^{(t)})
\end{equation}
The routing state (i.e. Retrieval SubTree in Section~\ref{sec:dst}) encapsulates structural information, enabling the $\mathrm{Router}$ to navigate through the document hierarchy. Each document $d_i$'s routing state $s_i^{(t)}$ is initialized with the corresponding retrieved chunks ${C_{re}^{(i)} \subseteq C_{re}}$ and updated by the $\mathrm{Router}$ actions of the previous round $\{\langle a_{ij}^{(t-1)}, p_{ij}^{(t-1)} \rangle\}_{j=1}^{n_i^{(t-1)}}$. Here, the operator $\otimes$ denotes content expansion:
\begin{equation}
  s_i^{(t)} = d_i \otimes
    \begin{cases}
      E_i^{(t-1)} & ,\  t > 1 \\
      C_{re}^{(i)} & ,\  t = 1
    \end{cases}
\end{equation}
\begin{equation}
  E_i^{(t)} = \{ p_{ij}^{(t)} \mid j \in \{1, ..., n_i^{(t)}\},\ a_{ij}^{(t)} = \mathsf{[EXP]} \}
\end{equation}
The routed chunk $c_{ro}^{(i)} \in C_{ro}$ is accumulated by aggregating the selected passages across all routing steps $t=\{1, \cdots, T_i\}$.
\begin{equation}
  c_{ro}^{(i)} = \bigoplus_{t=1}^{T_i} A_i^{(t)}
\end{equation}
\begin{equation}
  A_i^{(t)} = \{ p_{ij}^{(t)} \mid j \in \{1, ..., n_i^{(t)}\},\ a_{ij}^{(t)} = \mathsf{[ANS]} \}
\end{equation}
where the operator $\oplus$ denotes passage concatenation.

\textbf{Read.} The $\mathrm{Reader}$ (typically an LLM) generates the final answer $a$, conditioning on both the input question $q$ and the routed passages $C_{ro}$.
\begin{equation}
  \label{eq:reader}
  a = \mathrm{Reader}(q, [c_{ro}^{(1)}, \cdots, c_{ro}^{(m)}])
\end{equation}

\subsection{Document Structure Representation}
\label{sec:dst}

While standard RAG frameworks process only flat content chunks, our approach preserves critical structural information through formal tree representations. To capture hierarchical relationships in documents, we define two types of nodes: (1) \textit{Structure Nodes} represent organizational hierarchy (e.g., headings), and (2) \textit{Content Nodes} contain substantive textual information (e.g., passages).

\textbf{Document Structure Tree.} A Document Structure Tree (DST) encodes the full document hierarchy, where each node is represented as:
\begin{equation}
  \label{eq:dst}
  \text{DST-node} = \langle \textit{id}, \textit{text}, \tau, \textit{parent}, \mathcal{C} \rangle
\end{equation}

Here $\tau \in \{\text{structure}, \text{content}\}$ denotes the node type, and $\mathcal{C}$ indicates the ordered set of child nodes. Each node is defined by a unique identifier (\textit{id}), associated \textit{text} content - either a heading title (for structure nodes) or passage text (for content nodes) - and a pointer to its \textit{parent} node (null for the root). The root node, always a structure node, corresponds to the document title.

\textbf{Retrieval Subtree.} A Retrieval SubTree (RST) is derived from the DST, designed to maintain stable retrieval scope while adaptively updating contextual focus. An RST consists of (1) all structure nodes (complete document hierarchy), and (2) selected content nodes (partial content coverage).

During inference, the RST is first initialized with the retrieved passages along with their content siblings, then iteratively updated by replacing them with previously unseen content nodes under a single router-selected heading, while preserving all structure nodes (see Algorithm~\ref{alg:node_light} in Appendix~\ref{sec:a1}). This constrained derivation strategy ensures stable RST size while dynamically refining the contextual focus.

\subsection{Routing Module}
\label{sec:routing_module}

As shown in Figure~\ref{fig:routing_module}, the routing module synergistically combines document tree structure with an LLM-based router, enabling structure-aware retrieval-augmented generation.


\textbf{Task Formulation.} We define document routing task as iterative navigation through a document structure tree, dynamically assembling fine-grained passage chunks with both content relevance and structural integrity. This process emerges through compositional application of three atomic actions at each step:
\begin{compactitem}
  \item $\mathsf{[ANS]}$: Select a visible \textit{content node} when its text directly answers the question;
  \item $\mathsf{[EXP]}$: Unfold a collapsed \textit{structure node} if its heading text or contextual position suggests potential relevance;
  \item $\mathsf{[REF]}$: Stop exploring the current subtree when no nodes satisfy $\mathsf{[ANS]}$ or $\mathsf{[EXP]}$ criteria.
\end{compactitem}


\textbf{Action Curation.} Standard RAG datasets consists of a question with a reference answer, without providing the intermediate routing trajectories. We propose an automatic method for curating routing actions \textit{solely} from the question, requiring no necessary access to the answer. Specifically, given a question $q$, we first retrieve top-$k$ passages via an off-the-shelf retriever, access their originating documents, and derive corresponding retrieval subtrees $S$. We condition an LLM respectively on each subtree $s \in S$, along with the question $q$ to generate single-turn routing actions $A$. Finally, the routing dataset cruated consists of $\langle q, s, A \rangle$ triples.

\textbf{Training.} The training paradigm focuses on equipping the model with fundamental decision-making capabilities through exposure to individual routing actions (as opposed to complete iterative procedures). We fine-tune an LLM on the curated routing dataset using the standard next-token-prediction objective under supervised-fine-tuning (SFT), where the cross-entropy loss $\mathcal{L}$ is computed only on the target output tokens. This approach provides the necessary components for multi-step exploration during inference. 
\begin{equation}
  \label{eq:train}
  \mathcal{L} = -\log{P(A|q, s)}
\end{equation}

We convert document hierarchy into LLM-understandable text representation. Specifically, the input retrieval subtree uses the newline-delimited "\texttt{id: text}" format, where each level of hierarchy is represented by an additional indentation unit preceding the node identifier. The output action follows the "\texttt{[ACTION] id: text\_prefix}" format to ensure semantic grounding to the original id-text binding.

\section{Experiments}

\begin{figure*}[t]
  \includegraphics[width=\textwidth]{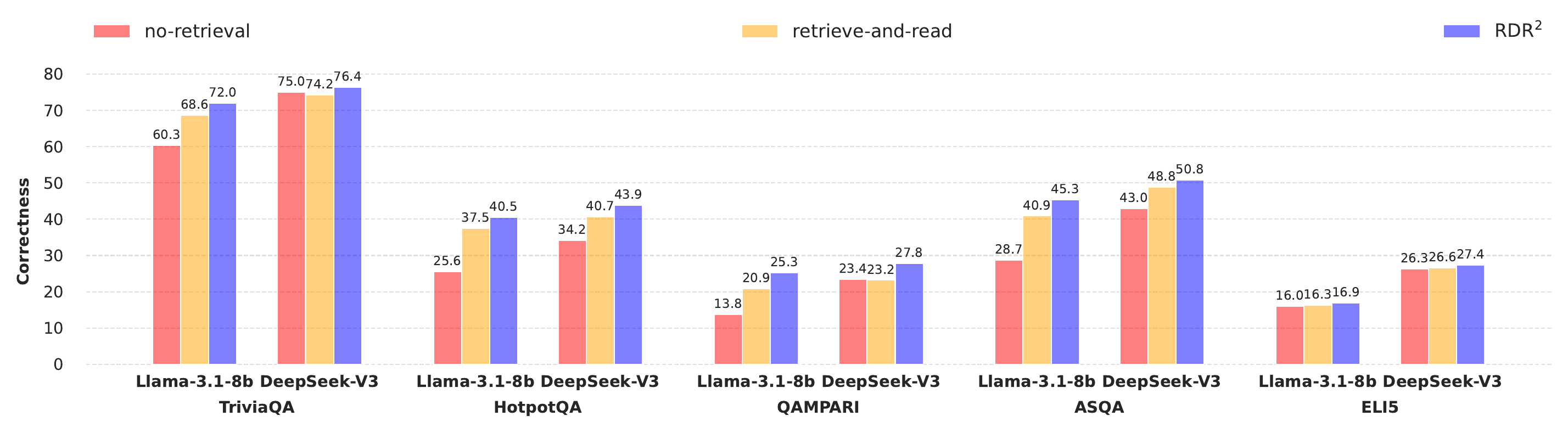}
  \caption{Comparison between RDR\textsuperscript{2} and baselines across all datasets with different readers. We report the primary correctness metric for each dataset: Exact Match for TriviaQA, HotpotQA and ASQA, F\textsubscript{1}-5 for QAMPARI and Claim Recall for ELI5.}
  \label{fig:main_baselines}
\end{figure*}

\subsection{Datasets and Metrics}

We evaluate RDR\textsuperscript{2} on five knowledge-intensive tasks across different QA formats. We follow previous works \citep{gao-etal-2023-enabling} to randomly sub-sample at most 1,000 examples from each dataset due to the experimental cost. Across all datasets, only the question field is used for both retrieval and generation, with Wikipedia consistently serving as the retrieval datastore.

\textbf{Short-form Generation.} \textbf{TriviaQA} \citep{joshi-etal-2017-triviaqa} consists of trivia questions, each calling for a single short answer. \textbf{HotpotQA} \citep{yang-etal-2018-hotpotqa} features Wikipedia-based question-anwer pairs requiring interleave retrieval and reasoning. For both datasets, we report \textit{EM} Recall and String \textit{F\textsubscript{1}}, following standard setups in \citet{asai2023self} and \citet{wang2025chain}.

\textbf{QAMPARI} \citep{amouyal-etal-2023-qampari} is a list-style QA dataset where answers comprise multiple factual short entities (avg. 13 instances) originated from diverse passages. We report \textit{F\textsubscript{1}-5}, \textit{precision} and \textit{recall-5}, where recall-5 is considered 100\% if at least five gold answers are covered, following ALCE \citep{gao-etal-2023-enabling} benchmark.

\textbf{Long-form Generation.} \textbf{ASQA} \citep{stelmakh-etal-2022-asqa} is a long-form factoid QA dataset featuring inherently ambiguous questions that requires RAG methods to reconcile diverse interpretations and produce coherent responses (avg. 65 words). We adopt the official metrics from the ASQA paper: \textit{EM} (Exact Match), \textit{ROUGE-L}, and \textit{Disambig-F\textsubscript{1}}.

\textbf{ELI5} \citep{fan-etal-2019-eli5} contains complex, diverse, open-ended questions derived from post titles in Reddit's "Explain Like I'm Five" forum, requiring systems to retrieve multiple documents and elaborate in-depth explanations (avg. 131 words). Following \citet{gao-etal-2023-enabling}, we use \textit{Claim Recall}, computed by checking whether the generated output entails reference sub-claims using an NLI model.

We additionally evaluate fluency and conciseness for long-form generation tasks. For fluency, we follow ALCE to use \textit{MAUVE} \citep{pillutla2021mauve} to assess distributional similarity between generated and ground-truth answers. For conciseness, we report response \textit{length} (in words), as longer outputs may artificially boost recall-type metrics (e.g., EM or Claim Recall).

\subsection{Baselines}
\label{sec:baselines}

We evaluate our framework against three categories of baselines: (1) \textit{No-Retrieval}: the reader directly answers questions using only its parametric knowledge, (2) \textit{Retrieve-and-Read}: the standard RAG pipeline with top-$k$ retrieved passages, and (3) \textit{Advanced RAG}: including methods based on proprietary LLMs: ASC and its variant ASC-F \citep{thirukovalluru-etal-2024-atomic}, as well as techniques fine-tuned on open-source LLMs: S{\small ELF}-R{\small EASONING} \citep{xia2025improving}, S{\small ELF}-RAG\footnote{We increased the $\mathsf{[Retrieve]}$ token probability by 0.2 to promote multi-turn retrieval for a fair comparison.} \citep{asai2023self}, O{\small PEN}-RAG\footnote{With only the 7B model publicly released, we fine-tuned the 13B variant using the official training script.} \citep{islam-etal-2024-open}, and F{\small RONT} \citep{huang-etal-2024-learning}.

\subsection{Experimental Settings}

For retrieval, we use the Wikipedia dump from \citet{karpukhin-etal-2020-dense}. We construct DSTs (defined in Section~\ref{sec:dst}) from the corresponding wiki pages, totaling 5.82M documents. Unless otherwise specified, we use the off-the-shelf Contriever-MS MARCO \citep{izacard2022unsupervised} as the retriever, with top-$5$ retrieved chunks for all retrieval-augmented methods.

We curate routing actions using Deepseek-v3 \citep{liu2024deepseek} following the procedure defined in Section~\ref{sec:routing_module} on ASQA training questions, resulting in 23,827 training samples 
and 500 test samples 
. The router is fine-tuned via LoRA \citep{hu2022lora} on Llama-3.1-8B-Instruct \citep{grattafiori2024llama} for 3.5 epochs, using LlamaFactory \citep{zheng-etal-2024-llamafactory} (see Appendix~\ref{sec:a1} for data curation details, Appendix~\ref{sec:a2} for training hyperparameters, and Appendix~\ref{sec:c} for router prompts).

Llama-2-13B-Chat \citep{touvron2023llama} and Llama-3.1-8B-Instruct \citep{grattafiori2024llama} are used as the open-source readers. To ensure fair comparison, we apply greedy decoding with model-specific maximum tokens, as significant inter-model length variations were observed (consistent with \citet{asai2023self}'s findings). For proprietary models including ChatGPT \citep{ouyang2022training} and Deepseek-v3 \citep{liu2024deepseek}, we set temperature=0.2 without length constraints, since their output lengths naturally align with the reference (see Appendix~\ref{sec:c} for reader prompts).

All experiments run on single NVIDIA A100-PCIE-40GB GPUs.


\section{Results and Analysis}

We first present overall results, then perform ablation studies to assess the contribution of each key component. Finally, we examine the framework's scalability under different test-time conditions and its robustness to various RAG component choices. A full case study is provided in Appendix~\ref{sec:d}.


\begin{table*}
  \centering
  \begin{tabular}{lccc|ccccc|ccc}
    \hline
      \multicolumn{1}{c}{} & \multicolumn{3}{c}{QAMPARI} & \multicolumn{5}{c}{ASQA} & \multicolumn{3}{c}{ELI5} \\
    \cmidrule(lr){2-4} \cmidrule(lr){5-9} \cmidrule(lr){10-12}
      & \textbf{F\textsubscript{1}-5} & \textbf{R-5} & \textbf{Pre} & \textbf{EM} & \textbf{D-F\textsubscript{1}} & \textbf{R-L} & \textbf{Mau} & \textbf{Len} & \textbf{Cla} & \textbf{Mau} & \textbf{Len} \\
    \hline
      \multicolumn{12}{c}{Reader based on ChatGPT} \\
      ASC-F & 18.8 & \textbf{45.0} & 13.4 & 45.0 & 31.9 & - & 41.3 & \textcolor{gray}{106.7} & 22.2 & \textbf{22.7} & \textcolor{gray}{172.7} \\
      ASC & 26.2 & 33.0 & 23.0 & 44.1 & 32.2 & - & 47.0 & \textcolor{gray}{101.2} & 21.4 & 21.3 & \textcolor{gray}{163.6} \\
    \cline{1-1}
      \textbf{RDR\textsuperscript{2}(Ours)} & \textbf{26.4} & 29.0 & \textbf{30.9} & \textbf{46.1} & \textbf{37.1} & \textbf{38.5} & \textbf{70.6} & \textcolor{gray}{49.1} & \textbf{23.3} & 14.4 & \textcolor{gray}{155.2} \\
    \hline
      \multicolumn{12}{c}{Reader fine-tuned on Llama-2-13B} \\
      S{\small ELF}-R{\small EASONING} & - & - & - & 35.2 & - & - & - & - & - & - & - \\
      S{\small ELF}-RAG\textsuperscript{*} & 6.5 & 9.0 & 6.4 & 37.5 & 27.5 & 39.2 & \textbf{77.7} & \textcolor{gray}{69.9} & 11.8 & \textbf{42.8} & \textcolor{gray}{81.9} \\
      O{\small PEN}-RAG\textsuperscript{*} & 2.5 & 3.9 & 2.3 & 39.9 & 24.0 & \textbf{40.4} & 17.2 & \textcolor{gray}{83.7} & 11.9 & 19.1 & \textcolor{gray}{129.2} \\
      F{\small RONT} & - & 11.9 & 22.6 & 41.5 & - & 38.6 & 76.1 & \textcolor{gray}{57.6} & 9.3 & 34.4 & \textcolor{gray}{75.1} \\
    \cline{1-1}
      \textbf{RDR\textsuperscript{2}(Ours)} & \textbf{23.2} & \textbf{24.3} & \textbf{25.0} & \textbf{41.7} & \textbf{31.6} & 39.2 & 61.2 & \textcolor{gray}{69.6} & \textbf{15.4} & 23.9 & \textcolor{gray}{148.3} \\
    \hline
  \end{tabular}
  \caption{\label{main_sotas}
  Comparison between RDR\textsuperscript{2} and other RAG methods on QAMPARI, ASQA and ELI5 wrt. corresponding metrics. 
  F\textsubscript{1}-5 is the harmonic mean of recall-5 (R-5) and precision (Pre), EM is Exact Match, D-F\textsubscript{1} is Disambig-F\textsubscript{1}, R-L is ROUGE-L, Mau is MAUVE, Cla is Claim Recall. 
  \textbf{Bold} indicates best results within each reader category. 
  \textcolor{gray}{Gray} denotes the word-level length (Len). 
  * marks the results from our reproduction.
  }
\end{table*}

\subsection{Main Results}

\textbf{Overall Performance.}
Figure~\ref{fig:main_baselines} evaluates the overall performance of RDR\textsuperscript{2} against two fundamental frameworks: \textit{no-retrieval} and \textit{Retrieve-and-Read}. Notably, in RDR\textsuperscript{2} only the router is finetuned on ASQA training questions (without answer supervision), while both retriever and reader remain off-the-shelf. TriviaQA, HotpotQA, QAMPARI and ELI5 serve as challenging generalization tests, being completely withheld from our router training.

\textbf{RDR\textsuperscript{2} continuously improves RAG performance.} 
With larger language models, standard \textit{Retrieve-and-Read} shows diminishing returns over \textit{no-retrieval}, suggesting their stronger parametric knowledge reduces reliance on retrieved content. While RDR\textsuperscript{2} also exhibits this scaling trend versus \textit{no-retrieval}, its improvement over \textit{Retrieve-and-Read} remains relatively stable across model scales, confirming the inherent value of document structure awareness in retrieval-augmented generation.

\textbf{RDR\textsuperscript{2} effectively generalizes to held-out datasets.}
While RDR\textsuperscript{2} maintains strong performance on QAMPARI comparable to its ASQA results, we observe limited gains on ELI5. This aligns with prior findings \citep{krishna-etal-2021-hurdles, jiang-etal-2023-active} on the intrinsic challenges of open-ended long-form QA, where the expansive space of potentially valid answers poses fundamental difficulties for retrieval-augmented approaches and their evaluation.

\textbf{Comparison with baselines.}
Table~\ref{main_sotas} compares RDR\textsuperscript{2} against cutting-edge RAG methods employing either proprietary LLMs (ChatGPT) or fine-tuned open-source Llama-2-13B variants as their backbone readers.

\textbf{RDR\textsuperscript{2} achieves new state-of-the-art results.} 
Across all three datasets - QAMPARI, ASQA and ELI5 - RDR\textsuperscript{2} consistently outperforms existing approaches, demonstrating strong generalization across diverse QA scenarios. Specifically:

It is noteworthy that among the compared methods based on open-source models, all require reader fine-tuning on carefully annotated question-answer pairs (some including training set of the downstream tasks), whereas our approach achieves superior performance using only readily available questions for router training, paired with an entirely off-the-shelf reader.

Furthermore, methods employing proprietary LLMs generate significantly longer responses (~2$\times$ the gold answer length on ASQA) to achieve high EM recall, while our approach attains better results with approximately 50\% shorter outputs. On QAMPARI, this verbosity leads to precision degradation, whereas our method maintains balanced precision-recall performance. These observations collectively validate our framework's enhanced efficiency in information delivery.

\subsection{Ablation Study}
\label{sec:ablation_study}

Table~\ref{ablation} presents comprehensive ablation studies analyzing three critical dimensions of our framework: \textit{pipeline architecture} (Section~\ref{sec:rdr2_framework}), \textit{router information} (Section~\ref{sec:dst}), and \textit{routing actions} (Section~\ref{sec:routing_module}). We evaluate both intermediate routed passages and final generated answers, measuring factual correctness through Exact Match (EM) and verbosity via word count (Len).

\begin{table}[h]
  \centering
  \begin{tabular}{lcccccc}
    \hline
      \multicolumn{1}{c}{} & \multicolumn{2}{c}{Passage} & \multicolumn{2}{c}{Answer} \\
    \cmidrule(lr){2-3} \cmidrule(lr){4-5}
      & \textbf{EM} & \textbf{Len} & \textbf{EM} & \textbf{Len} \\
    \hline
      \textbf{RDR\textsuperscript{2}(Ours)} & \underline{57.3} & \textcolor{gray}{104.2} & \textbf{45.3} & \textcolor{gray}{71.3} \\
    \cline{1-1}
      \hspace{1em} w/o router & 51.7 & \textcolor{gray}{100.0} & 40.9 & \textcolor{gray}{69.2} \\
    \cline{1-1}
      \hspace{1em} w/o structure & 49.8 & \textcolor{gray}{67.5} & 41.3 & \textcolor{gray}{71.0} \\
      \hspace{1em} w/o similarity & 54.8 & \textcolor{gray}{100.9} & \underline{43.9} & \textcolor{gray}{72.3} \\
        \hspace{2em} w/o content & 54.2 & \textcolor{gray}{93.9} & 43.7 & \textcolor{gray}{70.0} \\
    \cline{1-1}
      \hspace{1em} w/o $\mathsf{[EXP]}$ & 52.9 & \textcolor{gray}{81.7} & 42.5 & \textcolor{gray}{71.9} \\
      \hspace{1em} w/o $\mathsf{[REF]}$ & \textbf{61.2} & \textcolor{gray}{176.3} & 42.9 & \textcolor{gray}{70.7} \\
    \hline
  \end{tabular}
  \caption{\label{ablation}
    Ablation Study on ASQA. 
    Ablated variants (w/o = without) are defined in Section~\ref{sec:ablation_study}. 
    We report Exact Match (EM) and word-level length (\textcolor{gray}{Len}) for passages and answers. 
    \textbf{Bold} and \underline{Underline} denote best and second best results, respectively. 
  }
\end{table}

\subsubsection{Pipeline Architecture}
\label{sec:pipeline_architecture}

\textbf{Ablating Router.} Removing the routing module (\textbf{w/o router}) reduces the RAG pipeline to standard \textit{Retrieve-and-Read} framework. Our full framework significantly improves factual recall (+5.6 EM) while maintaining comparable passage length (104.2 vs. 100.0), demonstrating enhanced informativeness without compromising conciseness. This improvement carries through to answer generation (+4.4 EM), demonstrating consistent gains across the entire RAG pipeline.

\subsubsection{Router Information}

The router processes two types of information: (1) \textit{structure} from document headings, and (2) \textit{similarity} from retrieved passages. We ablate each component:

\textbf{Ablating Structure.} The \textbf{w/o structure} variant discards document hierarchy and use only retrieved passages\footnote{To ensure fair comparison, we reconstruct content at the node level to avoid information loss from chunk truncation.}, where the router simply accepts or refuses individual passages. We observe significant drops in both passage retrieval (-7.5 EM) and answer generation (-4.0 EM) versus the full framework, confirming structural cues provide critical gains. Compared to \textit{w/o router}, this ablation yields less informative passages (-1.9 EM) but better answers (+0.4 EM), showing structural awareness enables more effective knowledge organization despite occasional over-filtering.

\textbf{Ablating Similarity.} The \textbf{w/o similarity} variant initializes the RST with content nodes under a \textit{random} heading instead of retrieved passage siblings. A stricter variant (\textbf{w/o content}) removes content nodes entirely, despite this configuration being completely unseen during training. Ablating similarity causes moderate performance drops (-2.5 EM passages, -1.4 EM answers), confirming that providing question-relevant content offers crucial guidance for structural understanding and document routing. The small gap between these variants (0.6 EM passages, 0.2 EM answers) demonstrates the router's trained structural reasoning generalizes to unseen document formats.

\subsubsection{Routing Actions}

We validate each atomic action's necessity for document routing:

\textbf{Ablating $\mathsf{[EXP]}$.} The router can only select or refuse among currently visible nodes, losing the ability to explore new subtrees (\textbf{w/o $\mathsf{[EXP]}$}). The noticeable declines versus full framework (-4.4 passage EM, -2.8 answer EM) confirms expansion is crucial for discovering content that can hardly be recalled by similarity alone. Yet still outperforms \textit{w/o router} (+1.2 passage EM, +1.6 answer EM), showing RAG can benefit from basic structure awareness.

\begin{figure*}[t]
  \includegraphics[width=0.48\linewidth]{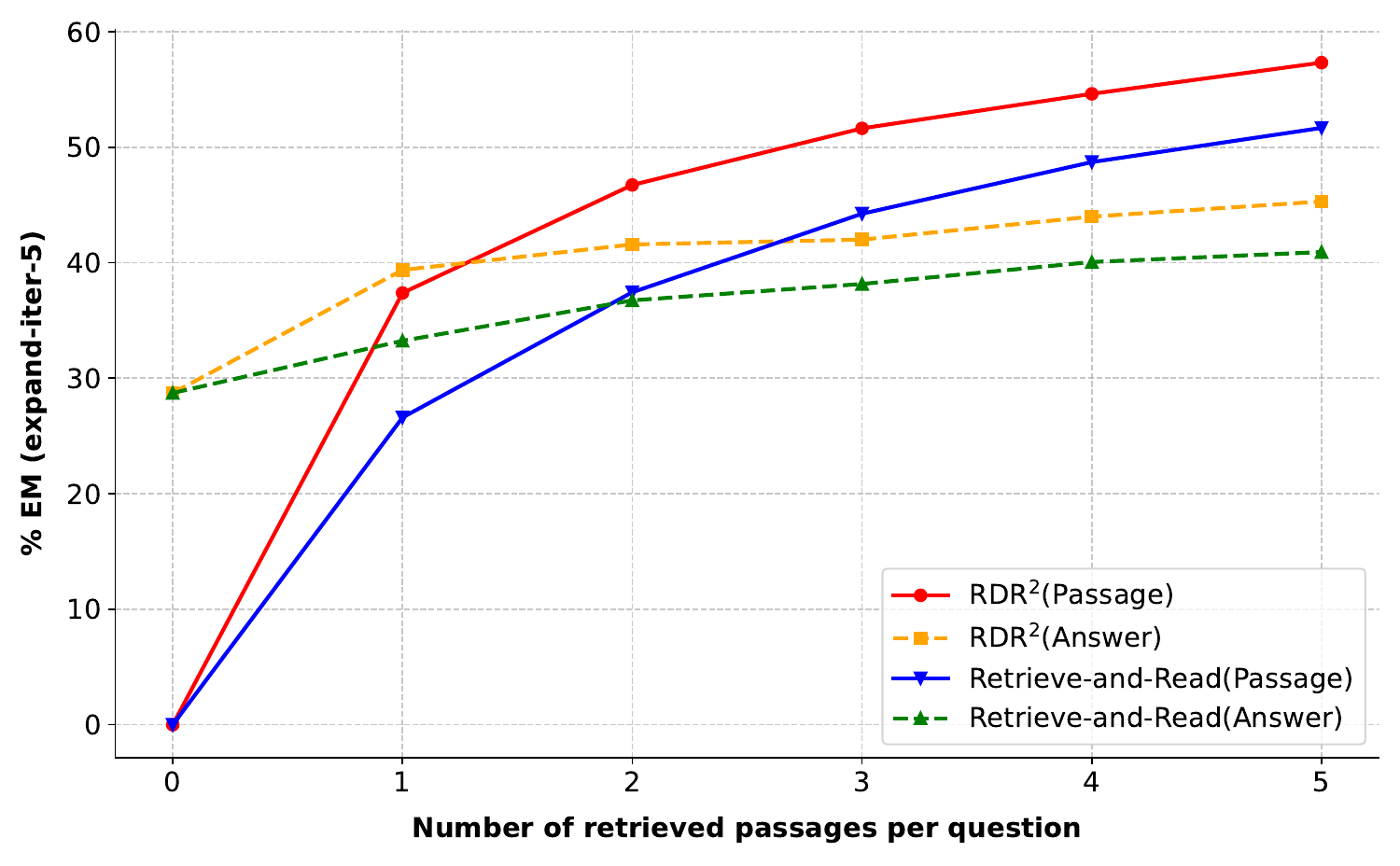} \hfill
  \includegraphics[width=0.48\linewidth]{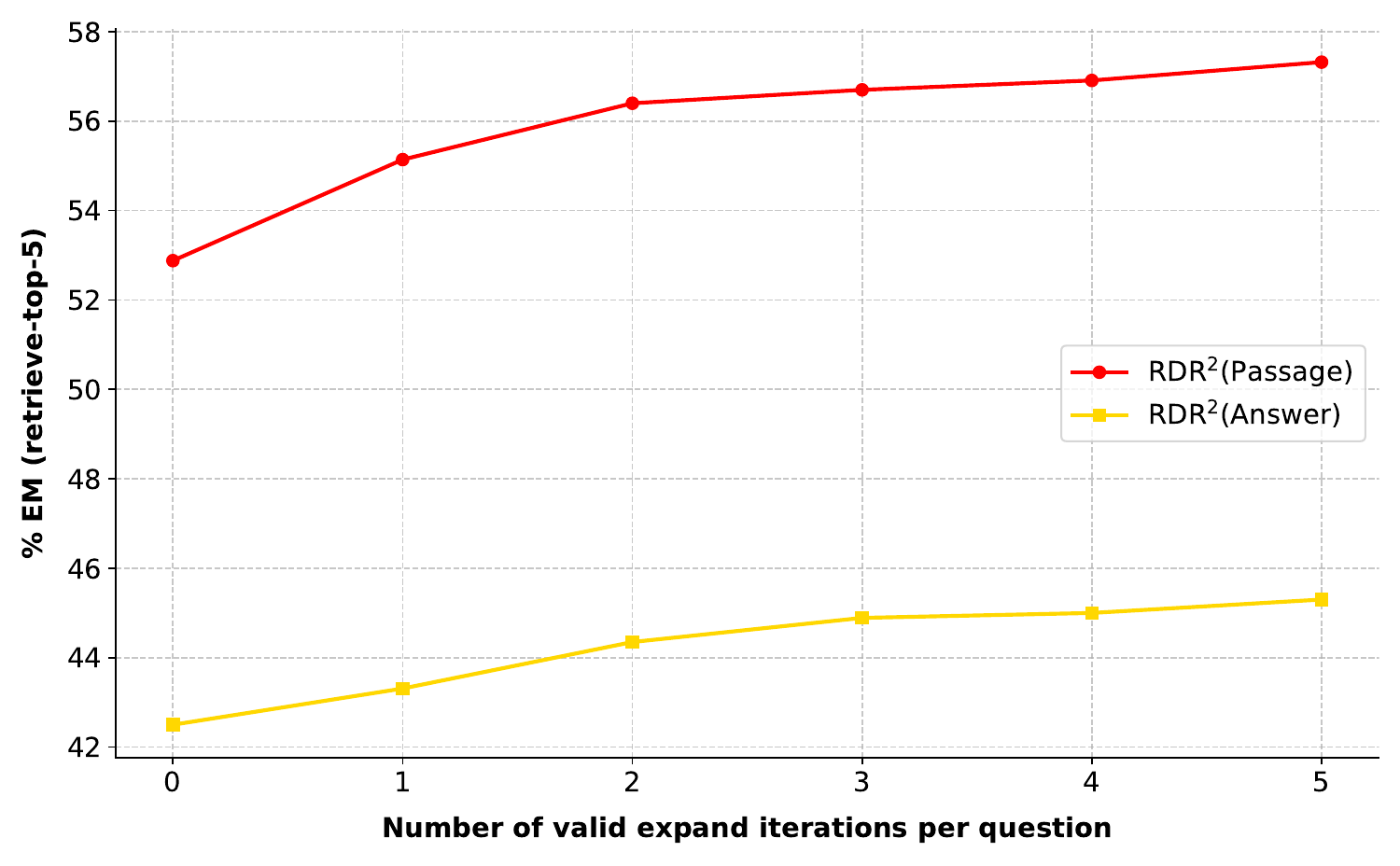}
  \caption {Scaling test-time compute on ASQA for RDR\textsuperscript{2} framework. \textit{Left}: \textit{top-$k$ scaling}. \textit{Right}: \textit{expand-$iter$ scaling}. Exact Match (EM) is reported from both passage/answer-aspect.}
  \label{fig:test_time_scaling}
\end{figure*}

\begin{figure*}[h]
  \includegraphics[width=0.48\linewidth]{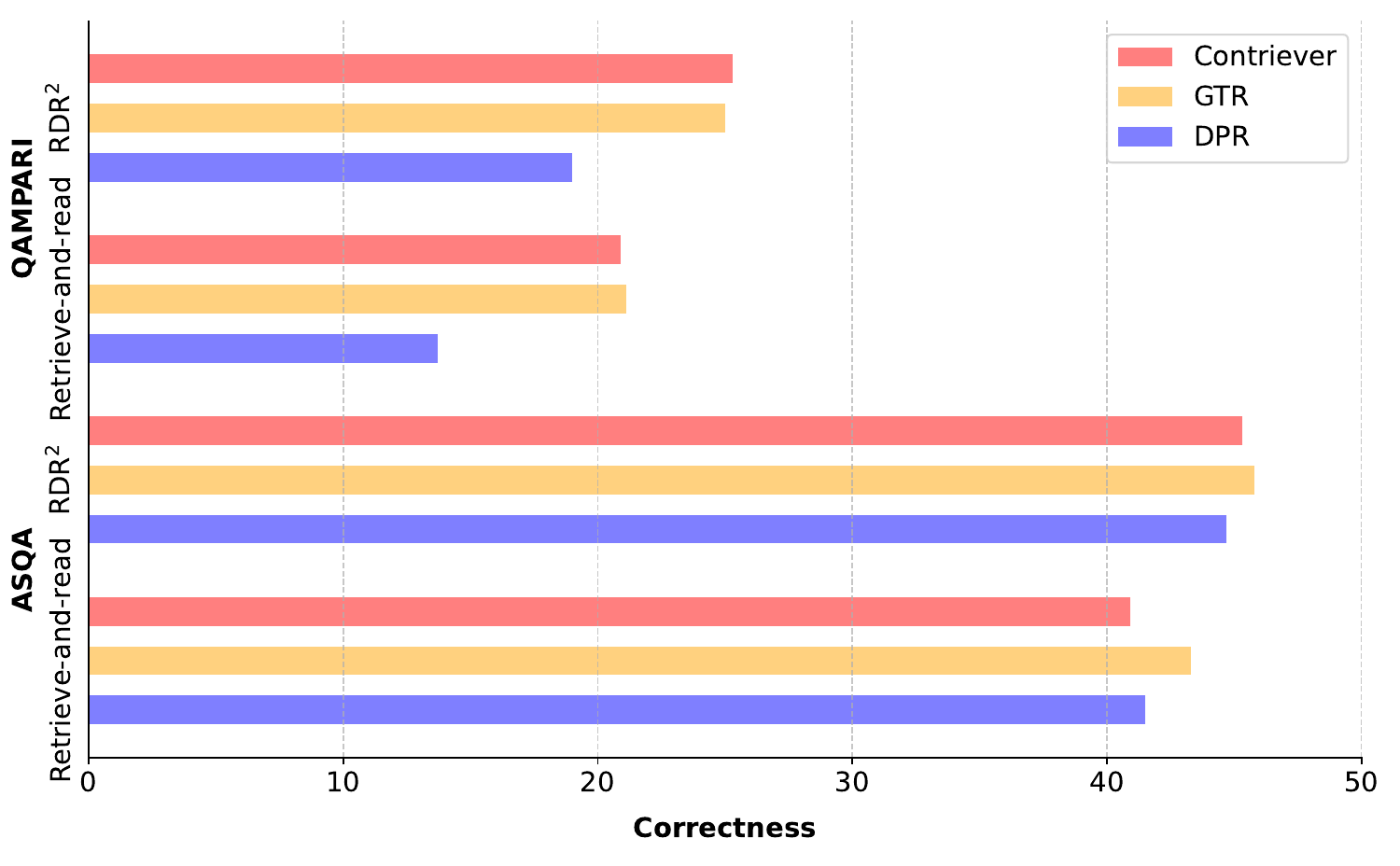} \hfill
  \includegraphics[width=0.48\linewidth]{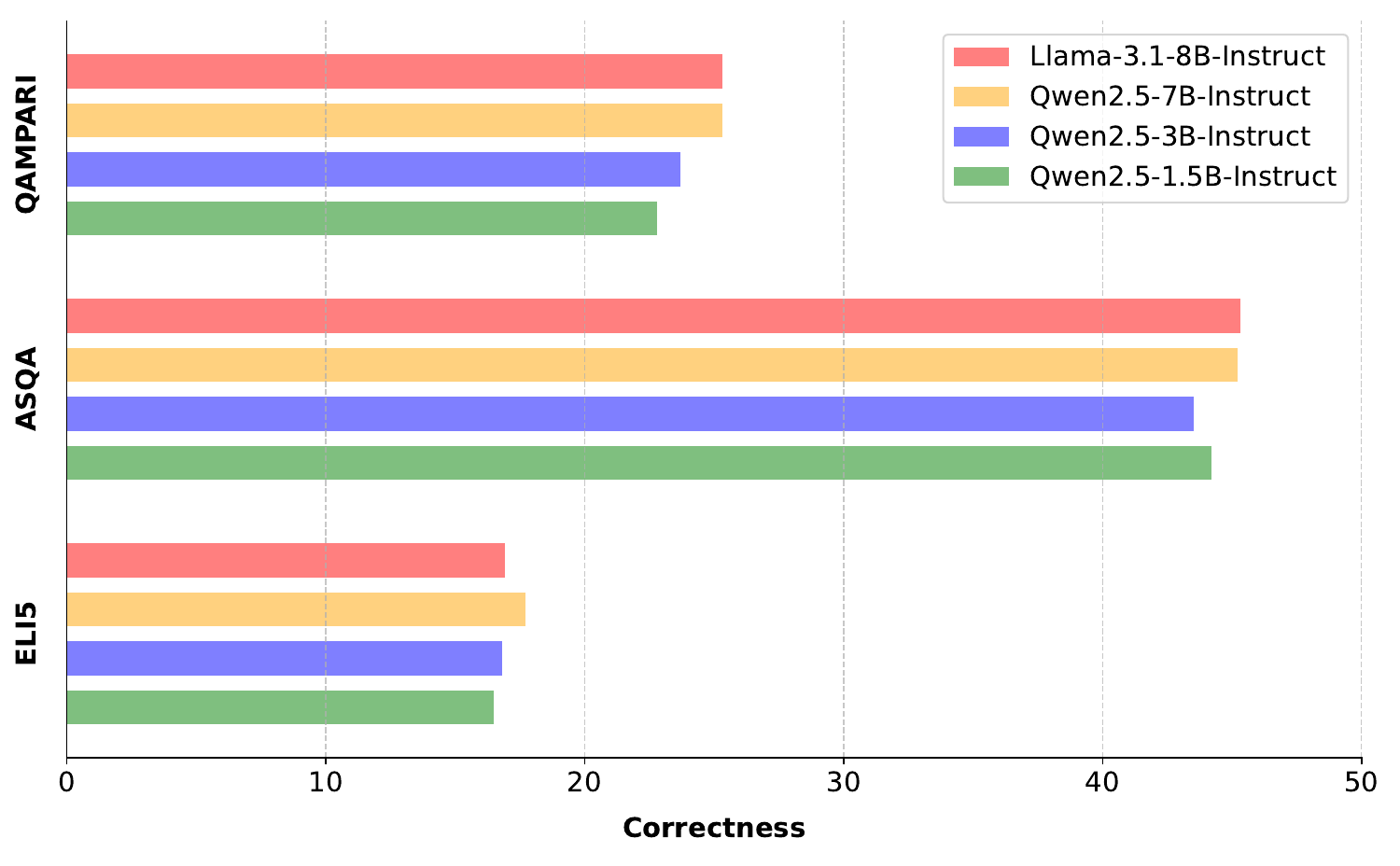}
  \caption {Robustness experiment across different datasets. \textit{Left}: retriever robustness. \textit{Right}: router robustness. Correctness metrics: F\textsubscript{1}-5 for QAMPARI, Exact Match for ASQA and Claim Recall for ELI5.}
  \label{fig:retriever_robustness}
\end{figure*}

\textbf{Ablating $\mathsf{[REF]}$.} The router must either answer or expand at least one node in each step, potentially forcing suboptimal choices (\textbf{w/o $\mathsf{[REF]}$}). Passage informativeness is substantially increased (+3.9 EM), yet its length doubled, introducing noise that ultimately harms answer quality (-2.4 EM), proving selective rejection is vital for concise knowledge organization.

\subsection{Test-time Scaling}

Inspired by OpenAI o1 \citep{jaech2024openai}'s observation, our framework enables dynamic test-time compute scaling without model weight updates. We investigate two scaling dimensions: (1) \textit{top-$k$ scaling} where we vary the number of retrieved passages $k \in [0, 5]$, and (2) \textit{expand-iter scaling} which controls document expansion iterations $iter \in [0, 5]$, With their impacts demonstrated in Figure~\ref{fig:test_time_scaling}.

\textbf{Top-$k$ Scaling.} As shown in Figure~\ref{fig:test_time_scaling} \textit{left}, increasing $k$ consistently improves both retrieval and answer correctness, as expanding the search space enhances the likelihood of capturing relevant documents. While standard \textit{Retrieve-and-Read} exhibits similar scaling trends, our framework maintains a consistent performance advantage. This suggests that structural awareness potentially enhances the benefits of retrieval test-time scaling.

\textbf{Expand-$iter$ Scaling.} As shown in Figure~\ref{fig:test_time_scaling} \textit{right}, increasing expansion iterations yields consistent improvements in both passage utility and answer quality. Our controlled expansion mechanism introduces a novel RAG scaling paradigm, offering adjustable trade-offs between performance and computational cost - particularly valuable for applications with varying latency-accuracy requirements.

\subsection{Robustness}

Figure~\ref{fig:main_baselines} demonstrates RDR\textsuperscript{2}'s robustness to diverse readers and held-out datasets. We further investigate the retrievers and routers compatibility.

\textbf{Retriever Robustness.} 
We use off-the-shelf GTR \citep{ni-etal-2022-large} and DPR \citep{karpukhin-etal-2020-dense} as the retriever. 
As shown in Figure~\ref{fig:retriever_robustness} \textit{left}, RDR\textsuperscript{2} maintains stable performance with different retrievers across datasets, while standard \textit{Retrieve-and-Read} exhibits performance fluctuations, empirically validates that explicit structure perception enhances RAG's robustness to component variations.

\textbf{Router Robustness.} We fine-tuned routers based on Qwen2.5-Instruct \citep{qwen2025qwen25technicalreport} series using the same protocol. As shown in Figure~\ref{fig:retriever_robustness} \textit{right}, experiments consistently validate our method's effectiveness across different model architectures and scales.



\section{Conclusion}

This work introduces RDR\textsuperscript{2}, a novel framework that explicitly incorporates document structure throughout the RAG process. Our approach dynamically navigates document structure trees using an LLM-based router, which jointly considers content relevance and hierarchical relationships to assemble optimal evidence. Comprehensive evaluations across five datasets demonstrate that document structure awareness brings significant and consistent gains to RAG systems, especially in scenarios requiring multi-document synthesis.

\section*{Limitations}

We acknowledge three key limitations of this work: (1) While our routing mechanism effectively navigates intra-document hierarchies, it processes each document independently. The document count is determined by the initial top-$k$ retrieval, potentially limiting inter-document knowledge integration. (2) The framework requires offline construction of Document Structure Trees (DSTs) for the entire datastore (approximately 20 minutes for Wikipedia in our experiment, with parallelization across 8 CPU cores). (3) The iterative routing process incurs computational overhead, which can be partially mitigated through controlled expansion iterations during inference.

\section*{Ethical Concerns}

This study focuses on improving knowledge acquisition and utilization in RAG systems through document structure awareness. All data, models, and APIs used in our experiments are sourced from publicly available platforms to ensure transparency and reproducibility. We strictly adhere to ethical guidelines throughout the research process, guaranteeing that our work poses no harm to individuals or groups. Furthermore, we commit to avoiding any form of deception or misuse of information in both methodology and application.



\bibliography{anthology,custom}

\appendix

\section{Implementation Details}
\label{sec:a}
\subsection{Dataset curation}
\label{sec:a1}

\begin{table}[hbt]
  \centering
  \begin{tabular}{ccccc}
  \toprule
     & ANS & EXP & REF & Total \\
  \midrule
    Train & 14,822 & 3,793 & 5,212 & 23,827 \\
    Test & 287 & 90 & 123 & 500 \\
  \bottomrule
  \end{tabular}
  \captionsetup{font=large}
  \caption{Routing Dataset.} \label{tab:route_dataset}
\end{table}

As shown in Table~\ref{tab:route_dataset}, the routing dataset consists of 23,827 training samples, including 14,822 $\mathsf{[ANS]}$ instances, 3,793 $\mathsf{[EXP]}$ instances, and 5,212 $\mathsf{[REF]}$ instances.


For the construction of the routing dataset, we begin by sampling queries from the ASQA training set and feed them into the retriever to obtain the top-$k$ relevant text chunks. Each retrieved chunk is then aligned with the original \textit{Document Structure Tree} by applying the Levenshtein Distance algorithm. Specifically, a sliding-window strategy with a stride of one character is employed to locate the most similar spans within the structure tree. Once the mapping between a retrieved chunk and its corresponding content node is established, we apply the \textit{RST Derivation Algorithm} (Algorithm~\ref{alg:node_light}) to systematically traverse the tree and retain all siblings, ancestors, and descendants of the mapped nodes that share the \textit{content type} attribute. This procedure yields the corresponding \textit{Retrieval Subtrees}. Finally, the DeepSeek-V3 API is leveraged to generate single-turn routing outputs conditioned on the given queries and their associated subtrees.


\begin{algorithm} [hbt]
    \caption{RST Derivation}
    \label{alg:node_light}
    \begin{algorithmic}[1]
        \Require $DST$, $Lighted\ nodes$
        \Function{LightNodes}{$Tree$, $Nodes$}
            \For{each $node \in Nodes$}
                \State $siblings \gets$ \Call{GetSiblings}{$Tree$, $node$}  \Comment{Acquiring necessary sibling nodes}
                \For{each $sibling \in siblings$}
                    \If{$sibling$.type = "content"}
                        \State $sibling$.lighted $\gets$ True
                    \EndIf
                \EndFor
                \State
                \State $current \gets node$
                \While{$current$.parent $\neq \emptyset$}
                    \State $current \gets current$.parent
                    \If{$current$.type = "structure"}
                        \State \textbf{break}  \Comment{Acquiring necessary upper ancestor nodes}
                    \EndIf
                    \State $current$.lighted $\gets$ True
                \EndWhile
                \State
                \For{each $sibling \in siblings$}
                    \If{$sibling$.type = "content"}
                        \State \Call{LightDescendants}{$Tree$, $sibling$}  \Comment{Acquiring necessary lower descendant nodes}
                    \EndIf
                \EndFor
            \EndFor
        \EndFunction
    \end{algorithmic}
\end{algorithm}

\subsection{Training details}
\label{sec:a2}

We choose Llama-3.1-8B-Instruct as the backbone of the routing model and employ LoRA for efficient fine-tuning. Specifically,  we set lora\_rank as 8, lora\_alpha as 16, gradient accumulated batch size as 8, learning rate as 1e-5 and epoch as 5. We also compare different training settings, as shown in Table~\ref{tab:train_settings}, and finally select the model based on instruct model with tag format prompt.

\begin{table*}[hbt]
  \centering
  \resizebox{\textwidth}{!}{
  \begin{tabular}{lcccc|ccccccc}
    \toprule
    \textbf{Prompt} & \textbf{Post-processing} & \textbf{Model} & \textbf{Train} & \textbf{Epoch} & \textbf{ANS-F1} & \textbf{EXP-PRE} & \textbf{REF-ACC} & \textbf{XPL-AVG} & \textbf{COL-RATE} \\
  \midrule
    enclose & √  & base & full & 7.0 & \textbf{86.3} & 46.9 & 82.9 & 0.6 & \textbf{0.0} \\
    enclose & √  & inst & full & 7.0 & 86.1 & 50.3 & 83.7 & 0.6 & \textbf{0.0} \\
    enclose & √  & inst & lora & 4.5 & 84.0 & 55.7 & 81.3 & \textbf{0.4} & \textbf{0.0} \\
    tag & √  & inst & lora & 3.5 & 83.0 & \textbf{57.1} & \textbf{87.0} & \textbf{0.4} & \textbf{0.0} \\
    tag & $\times$ & inst & lora & 5.0 & 84.4 & 51.1 & 77.2 & 7.6 & 0.4 \\
  \bottomrule
  \end{tabular}
  }
  \captionsetup{font=large}
  \caption{Comparison of different training settings. We evaluate performance of fine-tuned routing models on the curated test set. Specifically, ANS-F1 denotes the f1 score of $\mathsf{[ANS]}$ action, EXP-PRE indicates the precision of $\mathsf{[EXP]}$ action, REF-ACC 
represents the accuracy of the $\mathsf{[REF]}$ action, XPL-AVG is the percentage of expelled output, and COL-RATE is the rate of collapsed output. Enclose and tag prompt represent the format of "[expand]" and "<expand></expand>", respectively.} \label{tab:train_settings}
\end{table*}

\begin{figure*}[hbt]
 \centering
 \begin{minipage}[b]{0.5\textwidth}
    \includegraphics[width=\textwidth]{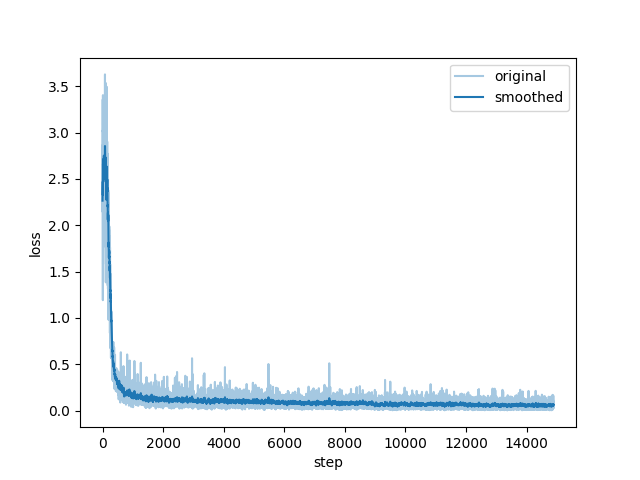}
    \captionsetup{font=large}
    \caption{Training loss curve of routing model.}\label{fig:router_training_loss}
 \end{minipage}
 \hfill
 \begin{minipage}[b]{0.47\textwidth}
    \includegraphics[width=\textwidth]{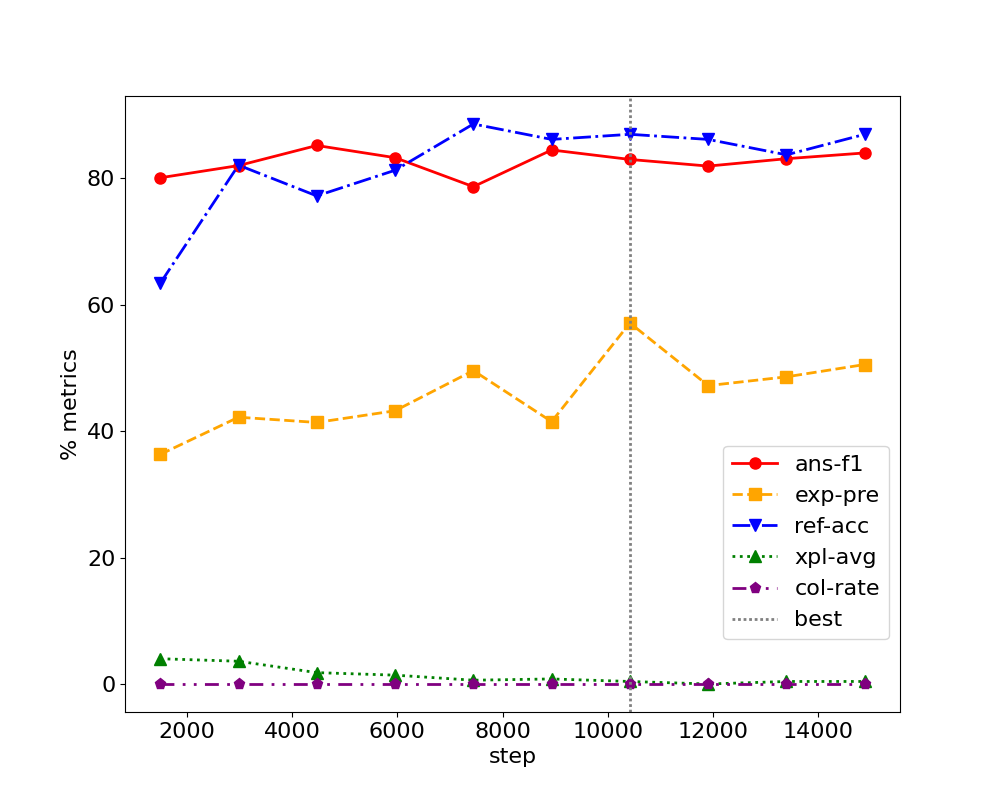}
    \captionsetup{font=large}
    \caption{Validation set performance of routing model.}\label{fig:router_dev_metrics}
 \end{minipage}
\end{figure*}

\section{More Experiments}
\label{sec:b}

\subsection{Main results}
\label{sec:b1}

As shown in Table~\ref{tab:full_main_asqa} and Table~\ref{tab:full_main_other}, we report full results of our main experiment. We can observe that:

1) With different backbone models, regardless of their openness or parameter scale, our framework consistently outperforms baseline methods across all evaluation metrics.

2) Compared to state-of-the-art approaches, our framework demonstrates superior performance on most metrics.

3) Furthermore, our framework significantly narrows the performance gap between open-source and proprietary models.

4) By learning document routing capabilities, our framework exhibits strong generalization ability on factual reasoning question answering tasks.

\begin{table*}
  \centering
  \resizebox{\textwidth}{!}{
  \begin{tabular*}{0.9\textwidth}{@{\extracolsep{\fill}}lccccc}
  \toprule
  \multicolumn{1}{c}{} & \multicolumn{5}{c}{ASQA} \\
  \cmidrule(lr){2-6}
    \textbf{Methods} & \textbf{EM(HIT)} & \textbf{D-F\textsubscript{1}} & \textbf{ROUGE} & \textbf{MAUVE} & \textbf{LEN} \\
  \midrule
  \multicolumn{6}{c}{Reader based on text-davinci-003} \\
    No-retrieval & 33.8 & 24.2 & 33.3 & - & - \\
    Retrieve-and-Read & 40.0 & 27.1 & 34.0 & - & - \\
    FLARE & \underline{41.3} & \underline{28.2} & \underline{34.3} & - & - \\
  \midrule
  \multicolumn{6}{c}{Reader based on ChatGPT} \\
    No-retrieval & 34.1(9.7) & 27.4 & 35.7 & 18.2 & 57.5 \\
    Retrieve-and-Read & 42.8(16.1) & 34.4 & 38.0 & 57.0 & 51.1 \\
    ASC-F & 45.0 & 31.9 & - & 41.3 & 106.7 \\
    ASC & 44.1 & 32.2 & - & 47.0 & 101.2 \\
    \textbf{RDR\textsuperscript{2}(Ours)} & \underline{46.1}(\underline{18.6}) & \underline{37.1} & \underline{38.5} & \underline{70.6} & 49.1 \\
  \midrule
  \multicolumn{6}{c}{Reader based on GPT-4o} \\
    No-retrieval & 41.4(13.7) & 33.9 & 36.2 & 23.3 & 58.8 \\
    Retrieve-and-Read & 47.0(19.1) & 36.5 & \underline{38.4} & 39.9 & 68.4 \\
    \textbf{RDR\textsuperscript{2}(Ours)} & \underline{48.2}(\underline{21.0}) & \underline{39.0} & \underline{38.4} & \underline{48.3} & 63.8 \\
  \midrule
  \multicolumn{6}{c}{Reader based on DeepSeek-V3} \\
    No-retrieval & 43.0(16.7) & 33.1 & 36.3 & 21.9 & 69.2 \\
    Retrieve-and-Read & 48.8(21.9) & 37.4 & 37.5 & 36.7 & 74.2 \\
    \textbf{RDR\textsuperscript{2}(Ours)} & \textbf{\underline{50.8}}(\textbf{\underline{23.2}}) & \textbf{\underline{39.8}} & \underline{37.8} & \underline{37.3} & 68.9 \\
  \midrule
  \multicolumn{6}{c}{Reader based on Llama-2-13b} \\
    No-retrieval & 24.7(6.5) & 19.3 & 35.1 & 13.4 & 65.9 \\
    Retrieve-and-Read & 36.5(13.5) & 26.9 & 39.2 & 31.4 & 61.2 \\
    S{\small ELF}-R{\small EASONING}(FT) & 35.2 & - & - & - & - \\
    S{\small ELF}-RAG(FT) & 31.7(8.4) & 26.4 & 37.0 & 71.6 & 27.0 \\
    S{\small ELF}-RAG(FT)\textsuperscript{*} & 37.5(14.9) & 27.5 & 39.2 & 77.7 & 69.9 \\
    O{\small PEN}-RAG(FT) & 36.3 & - & 38.1 & \textbf{\underline{80.0}} & - \\
    O{\small PEN}-RAG(FT)\textsuperscript{*} & 39.9(14.5) & 24.0 & \textbf{\underline{40.4}} & 17.2 & 83.7 \\
    F{\small RONT} & 41.5 & - & 38.6 & 76.1 & 57.6 \\
    \textbf{RDR\textsuperscript{2}(Ours)} & \underline{41.7}(\underline{16.9}) & \underline{31.6} & 39.2 & 61.2 & 69.6 \\
  \midrule
  \multicolumn{6}{c}{Reader based on Llama-3.1-8b} \\
    No-retrieval & 28.7(7.5) & 22.0 & 34.7 & 40.7 & 65.2 \\
    Retrieve-and-Read & 40.9(15.9) & 30.9 & 37.9 & 73.6 & 69.2 \\
    \textbf{RDR\textsuperscript{2}(Ours)} & \underline{45.3}(\underline{18.7}) & \underline{34.9} & \underline{38.2} & \underline{79.2} & 71.3 \\
  \bottomrule
  \end{tabular*}
  }
  \captionsetup{font=large}
  \caption{Main results on ASQA dataset. We report full results of different API and open-source models, together with results of no-retrieval and retrieve-and-read baselines. \textbf{Bold} and \underline{Underline} denote the best overall and in-category results, respectively. FT refers to methods finetuned on the corresponding training set. * marks the results from our reproduction.} \label{tab:full_main_asqa}
\end{table*}

\begin{table*}
  \centering
  \resizebox{\textwidth}{!}{
  \begin{tabular*}{0.9\textwidth}{@{\extracolsep{\fill}}lcccccc}
  \toprule
  \multicolumn{1}{c}{} & \multicolumn{3}{c}{QAMPARI} & \multicolumn{3}{c}{ELI5} \\
  \cmidrule(lr){2-4} \cmidrule(lr){5-7}
    Methods & \textbf{F\textsubscript{1}\text{-5}(F\textsubscript{1})} & \textbf{REC-5(REC)} & \textbf{PRE} & \textbf{CLAIM} & \textbf{MAUVE} & \textbf{LEN} \\
  \midrule
  \multicolumn{7}{c}{Reader based on ChatGPT} \\
    No-retrieval & 17.7(12.5) & 18.5(10.6) & 20.8 & \underline{23.6} & 14.4 & 145.3 \\
    Retrieve-and-Read & 22.2(15.7) & 22.1(13.4) & 27.9 & 22.3 & 12.2 & 141.5 \\
    ASC-F & 18.8(15.7) & \textbf{\underline{45.0}}(\textbf{\underline{29.8}}) & 13.4 & 22.2 & \underline{22.7} & 172.7 \\
    ASC & 26.2(19.5) & 33.0(20.5) & 23.0 & 21.4 & 21.3 & 163.6 \\
    \textbf{RDR\textsuperscript{2}(Ours)} & \underline{26.4}(\underline{19.8}) & 29.0(18.7) & \underline{30.9} & 23.3 & 14.4 & 155.2 \\
  \midrule
  \multicolumn{7}{c}{Reader based on GPT-4o} \\
    No-retrieval & 26.0(19.3) & 28.7(17.6) & 30.5 & \underline{26.5} & \underline{20.2} & 158.4 \\
    Retrieve-and-Read & 23.7(17.0) & 23.2(14.7) & 31.0 & 24.5 & 17.9 & 154.7 \\
    \textbf{RDR\textsuperscript{2}(Ours)} & \textbf{\underline{28.4}}(\underline{21.4}) & \underline{30.8}(\underline{20.3}) & \textbf{\underline{34.8}} & 25.3 & 16.9 & 165.3 \\
  \midrule
  \multicolumn{7}{c}{Reader based on DeepSeek-V3} \\
    No-retrieval & 23.4(18.4) & 28.7(18.8) & 23.3 & 26.3 & \underline{15.6} & 137.4 \\
    Retrieve-and-Read & 23.2(17.1) & 24.6(15.7) & 27.3 & 26.6 & 14.9 & 132.4 \\
    \textbf{RDR\textsuperscript{2}(Ours)} & \underline{27.8}(\textbf{\underline{21.7}}) & \underline{32.1}(\underline{21.6}) & \underline{31.1} & \textbf{\underline{27.4}} & 13.2 & 152.3 \\
  \midrule
  \multicolumn{7}{c}{Reader based on Llama-2-13b} \\
    No-retrieval & 14.9(10.3) & 16.4(9.0) & 14.3 & 14.7 & 21.9 & 140.2 \\
    Retrieve-and-Read & 21.0(14.7) & 22.0(12.9) & 21.6 & 14.9 & 20.8 & 141.2 \\
    S{\small ELF}-RAG & - & 1.9 & 1.3 & 6.1 & - & - \\
    S{\small ELF}-RAG\textsuperscript{*} & 6.5(4.9) & 9.0(5.5) & 6.4 & 11.8 & \textbf{\underline{42.8}} & 81.9 \\
    O{\small PEN}-RAG\textsuperscript{*} & 2.5(1.9) & 3.9(2.3) & 2.3 & 11.9 & 19.1 & 129.2 \\
    F{\small RONT} & - & 11.9 & 22.6 & 9.3 & 34.4 & 75.1 \\
    \textbf{RDR\textsuperscript{2}(Ours)} & \underline{23.2}(\underline{16.7}) & \underline{24.3}(\underline{14.9}) & \underline{25.0} & \underline{15.4} & 23.9 & 148.3 \\
  \midrule
  \multicolumn{7}{c}{Reader based on Llama-3.1-8b} \\
    No-retrieval & 13.8(10.3) & 19.3(11.1) & 13.1 & 16.0 & 18.8 & 139.5 \\
    Retrieve-and-Read & 20.9(15.1) & 23.6(14.3) & 22.9 & 16.3 & \underline{21.6} & 141.9 \\
    \textbf{RDR\textsuperscript{2}(Ours)} & \underline{25.3}(\underline{19.5}) & \underline{32.3}(\underline{21.1}) & \underline{25.7} & \underline{16.9} & 20.3 & 141.6 \\
  \bottomrule
  \end{tabular*}
  }
  \captionsetup{font=large}
  \caption{Main results on QAMPARI and ELI5 datasets. We report full results of different API and open-source models, together with results of no-retrieval and retrieve-and-read baselines. \textbf{Bold} and \underline{Underline} denote the best overall and in-category results, respectively. * marks the results from our reproduction.} \label{tab:full_main_other}
\end{table*}

\subsection{Ablation Study}
\label{sec:b2}

Full results of the ablation study are shown in Table~\ref{tab:full_ablation}. To evaluate the end-to-end ranking correctness of the retrieval process, we propose the \textbf{Inverse Information Rank Score} ${\rm Score}_{psg}$. Given the set of retrieved passages $C_{re} = \{c_{re}^{(i)}\}_{i=1}^{\rm top_k}$ and the set of reference short answers $A$, the score is defined as follows. 
 
\begin{align}
    & {\rm Score}_{psg} = \frac{\sum_{i=1:|C_{re}|}{\frac{1}{i} \cdot {\rm EM}(c_{re}^{(i)}, A)}}{|C_{re}|}
\end{align}

This metric models the gain of correctness information with a position-based decay, which aligns with the tendency of both retrieval and generation modules to favor top-ranked results.

\begin{table*}
  \small
  \centering
  \resizebox{\textwidth}{!}{
  \begin{tabular*}{0.9\textwidth}{@{\extracolsep{\fill}}lcccccc}
  \toprule
    Method & \textbf{P-EM(P-HIT)} & \textbf{P-SCORE} & \textbf{P-LEN} & \textbf{EM(HIT)} & \textbf{D-F\textsubscript{1}} & \textbf{LEN} \\
  \midrule
    \textbf{RDR\textsuperscript{2}(Ours)} & 57.3(34.2) & 12.7 & 104.2 & \textbf{45.3}(\textbf{18.7}) & \textbf{38.2} & 71.3 \\
    \cline{1-1}
        \hspace{1em} w/o router & 51.7(28.3) & 10.2 & 100.0 & 40.9(15.9) & 30.9 & 69.2 \\
    \cline{1-1}
        \hspace{1em} w/o structure & 49.8(28.0) & 10.6 & 67.5 & 41.3(15.0) & 32.5 & 71.0 \\
        \hspace{1em} w/o similarity & 54.8(32.7) & 11.8 & 100.9 & 43.9(17.8) & 33.2 & 72.3 \\
            \hspace{2em} w/o content & 54.2(31.3) & 11.7 & 93.9 & 43.7(18.0) & 34.0 & 70.0 \\
    \cline{1-1}
        \hspace{1em} w/o [expand] & 52.9(30.8) & 11.5 & 81.7 & 42.5(16.1) & 32.5 & 71.9 \\
        \hspace{1em} w/o [refuse] & \textbf{61.2}(\textbf{37.0}) & \textbf{13.4} & 176.3 & 42.9(16.4) & 32.8 & 70.7 \\
  \bottomrule
  \end{tabular*}
  }
  \captionsetup{font=large}
  \caption{Ablation results of RDR\textsuperscript{2}(ASQA).} \label{tab:full_ablation}
\end{table*}

\subsection{Test-time Scaling}
\label{sec:b3}

We report statistics of test-time scaling in Table~\ref{tab:top_k_apendix} and Table~\ref{tab:expand_iter_apendix}, including Top-\textit{k} and Expand-\textit{iter} scaling.

\subsection{Chunking Comparison}
\label{sec:b4}

We conduct a comparative experiment with Meta-Chunking \citep{zhao2024meta} on ASQA. Specifically, we applied Meta-Chunking to the retrieved documents with \textit{target\_size} = 100 to ensure comparable chunking sizes and \textit{threshold} = 0 for perplexity chunking. For a fair comparison, we carefully adapted Meta-Chunking to our setting, with an edit-distance constraint calculated against the original chunking to mitigate potential mismatch introduced by different retrievers. The results are shown in the Table~\ref{tab:meta_chunking} (with Llama3.1-8B-Instruct based router and reader).

\begin{table*}
  \centering
  \resizebox{\textwidth}{!}{  
  \begin{tabular}{lcccccccc}
  \toprule
    \textbf{Methods} & \textbf{EM(HIT)} & \textbf{D-F\textsubscript{1}} & \textbf{ROUGE} & \textbf{MAUVE} & \textbf{LEN} \\
  \midrule
    \textbf{Retrieve-and-Read}\\
    \cline{1-1}
        \hspace{1em} top-0 & 28.7(7.5) & 22.0 & 34.7 & 40.7 & 65.2 \\
        \hspace{1em} top-1 & 33.2(11.8) & 25.4 & 35.1 & 68.1 & 65.2 \\
        \hspace{1em} top-2 & 36.7(13.2) & 28.2 &36.8 & 70.2 & 66.6 \\
        \hspace{1em} top-3 & 38.2(14.0) & 29.5 & 37.3 & 75.8 & 69.7 \\
        \hspace{1em} top-4 & 40.1(14.6) & 30.7 & 37.7 & 74.0 & 70.5 \\
        \hspace{1em} top-5 & 40.9(15.9) & 30.9 & 37.9 & 73.6 & 69.2 \\
    \cline{1-6}
    \textbf{RDR\textsuperscript{2}(Ours)}\\
    \cline{1-1}
        \hspace{1em} top-0 & 28.7(7.5) & 22.0 & 34.7 & 40.7 & 65.2 \\
        \hspace{1em} top-1 & 39.4(15.1) & 30.2 & 36.0 & 69.1 & 70.5 \\
        \hspace{1em} top-2 & 41.6(16.5) & 32.6 & 37.7 & 71.8 & 69.6 \\
        \hspace{1em} top-3 & 42.0(16.5) & 32.9 & 37.8 & 77.7 & 68.6 \\
        \hspace{1em} top-4 & 44.0(17.1) & 33.9 & 38.2 & 76.2 & 69.4 \\
        \hspace{1em} top-5 & 45.3(18.7) & 34.9 & 38.2 & 79.2 & 71.3 \\
  \bottomrule
  \end{tabular}
  }
  \captionsetup{font=large}
  \caption{Statistics of Top-\textit{k} scaling.} \label{tab:top_k_apendix}
\end{table*}

\begin{table*}
  \centering
  \resizebox{\textwidth}{!}{  
  \begin{tabular}{lccc|ccccc}
  \toprule
    \textbf{Methods} & \textbf{P-EM(P-HIT)} & \textbf{P-SCORE}	& \textbf{P-LEN}
 & \textbf{EM(HIT)} & \textbf{D-F\textsubscript{1}} & \textbf{ROUGE} & \textbf{MAUVE} & \textbf{LEN} \\
  \midrule
    \textbf{RDR\textsuperscript{2}(Ours)}\\
    \cline{1-1}
        \hspace{1em} iter-0 & 52.9(30.8) & 11.5 & 81.7 & 42.5(16.1) & 32.5 & 37.7 & 76.7 & 71.9 \\
        \hspace{1em} iter-1 & 55.1(32.7) & 12.1 & 95.5 & 43.3(17.1) & 33.3 & 37.9 & 75.9 & 72.0 \\
        \hspace{1em} iter-2 & 56.4(33.2) & 12.5 & 95.4 & 44.3(18.9) & 34.5 & 38.1 & 78.7 & 69.3 \\
        \hspace{1em} iter-3 & 56.7(33.5) & 12.6 & 98.9 & 44.9(19.6) & 35.1 & 38.2 & 78.9 & 68.9 \\
        \hspace{1em} iter-4 & 56.9(33.7) & 12.6 & 100.1 & 45.0(18.6) & 34.8 & 38.2 & 76.0 & 71.1 \\
        \hspace{1em} iter-5 & 57.3(34.2) & 12.7 & 104.2 & 45.3(18.7) & 34.9 & 38.2 & 79.2 & 71.3 \\
  \bottomrule
  \end{tabular}
  }
  \captionsetup{font=large}
  \caption{Statistics of Expand-\textit{iter} scaling.} \label{tab:expand_iter_apendix}
\end{table*}

\begin{table*}
  \centering
  \resizebox{\textwidth}{!}{
  \begin{tabular*}{1.0\textwidth}{@{\extracolsep{\fill}}lcccccc}
  \toprule
  \multicolumn{1}{c}{} & \multicolumn{1}{c}{} & \multicolumn{5}{c}{ASQA} \\
  \cmidrule(lr){3-7}
    \textbf{Methods} & \textbf{Chunking} & \textbf{EM(HIT)} & \textbf{D-F\textsubscript{1}} & \textbf{ROUGE} & \textbf{MAUVE} & \textbf{LEN} \\
  \midrule
  No-retrieval & - & 28.7(7.5) & 22.0 & 34.7 & 40.7 & 65.2 \\
  Retrieve-and-Read & Fix-chunking & 40.9(15.9) & 30.9 & 37.9 & 73.6 & 69.2 \\
  Retrieve-and-Read & Meta-Chunking (MSP) & 41.3(16.4) & 30.0 & 37.0 & 76.1 & 74.3 \\
  Retrieve-and-Read & Meta-Chunking (PPL) & 40.8(15.2) & 29.4 & 37.1 & 70.6 & 75.4 \\
  \textbf{RDR\textsuperscript{2}} & - & \textbf{45.3(18.7)} & \textbf{34.9} & \textbf{38.2} & \textbf{79.2} & 71.3 \\
  \bottomrule
  \end{tabular*}
  }
  \captionsetup{font=large}
  \caption{Comparison with Meta-Chunking on ASQA dataset.} \label{tab:meta_chunking}
\end{table*}

\subsection{Short-form QA Performance}
\label{sec:b5}

We conduct experiments on two short-form datasets, HotpotQA and TriviaQA. As shown in Table~\ref{tab:short_form} ,compared to standard RAG, our method achieves consistent improvements: +8.0\% EM on HotpotQA and +5.0\% EM on TriviaQA. For reference, the correctness improvements on ASQA, QAMPARI and ELI5 are +10.8\%, +21.1\% and +3.7\% respectively. These improvements can be attributed to different underlying mechanisms. For HotpotQA, document structures facilitate multi-hop reasoning by explicitly modeling relationships between passages. For TriviaQA, where answers often rely on single passages, the gain likely stems from better context organization via structural awareness.

\begin{table*}
  \centering
  \resizebox{\textwidth}{!}{
  \begin{tabular*}{1.0\textwidth}{@{\extracolsep{\fill}}lccccc}
  \toprule
  \multicolumn{1}{c}{} & \multicolumn{1}{c}{} & \multicolumn{2}{c}{HotpotQA} & \multicolumn{2}{c}{TriviaQA} \\
  \cmidrule(lr){3-4} \cmidrule(lr){5-6}
  \textbf{Method} & \textbf{Router (SFT)} & \textbf{EM} & \textbf{F1} & \textbf{EM} & \textbf{F1} \\
  \midrule
  No-Retrieval & - & 25.6 & 28.2 & 60.3 & 60.1 \\
  Retrieve-and-Read & - & 37.5 & 41.8 & 68.6 & 69.0 \\
  \textbf{RDR\textsuperscript{2}} & Llama-3.1-8B-Instruct & \textbf{40.5} & 44.3 & \textbf{72.0} & \textbf{72.6} \\
  \textbf{RDR\textsuperscript{2}} & Qwen2.5-7B-Instruct & 39.8 & \textbf{44.7} & 69.9 & 70.9 \\
  \bottomrule
  \end{tabular*}
  }
  \captionsetup{font=large}
  \caption{Results on HotpotQA and TriviaQA datasets.} \label{tab:short_form}
\end{table*}

\subsection{Router Generalization Experiments}
\label{sec:b6}

As shown in Table\ref{tab:router_generalization}, we finetuned Qwen2.5-7B-Instruct as the router, observing comparable performance within each datasets. Moreover, experiments with routers finetuned on Qwen2.5 series (1.5B/3B/7B) consistently validate our method's effectiveness across different model scales.

\begin{table*}
  \centering
  \small  
  \setlength{\tabcolsep}{3pt}  
  \resizebox{\textwidth}{!}{
  \begin{tabular*}{1.0\textwidth}{@{\extracolsep{\fill}}lcccccccccc}
  \toprule
  \multicolumn{1}{c}{} & \multicolumn{1}{c}{} & \multicolumn{4}{c}{ASQA} & \multicolumn{3}{c}{QAMPARI} & \multicolumn{2}{c}{ELI5} \\
  \cmidrule(lr){3-6} \cmidrule(lr){7-9} \cmidrule(lr){10-11}
  \textbf{Router (SFT)} & \textbf{Reader} & \textbf{EM(HIT)} & \textbf{D-F\textsubscript{1}} & \textbf{R-L} & \textbf{Mau} & \textbf{F1-5(F1)} & \textbf{Rec-5(Rec)} & \textbf{Pre} & \textbf{Cla} & \textbf{Mau} \\
  \midrule
  Llama-3.1-8B-Inst & Llama-3.1-8B-Inst & \textbf{45.3}(18.7) & 34.9 & 38.2 & \textbf{79.2} & \textbf{25.3}(\textbf{19.5}) & \textbf{32.3}(\textbf{21.1}) & \textbf{25.7} & 16.9 & 20.3 \\
  Qwen2.5-7B-Inst & Llama-3.1-8B-Inst & 45.2(\textbf{18.9}) & \textbf{35.1} & \textbf{38.7} & 75.4 & \textbf{25.3}(19.3) & 31.2(19.9) & \textbf{25.7} & \textbf{17.7} & \textbf{24.5} \\
  Qwen2.5-3B-Inst & Llama-3.1-8B-Inst & 43.5(18.6) & 33.3 & 38.0 & 73.7 & 23.7(17.8) & 28.7(18.1) & 24.7 & 16.8 & 23.1 \\
  Qwen2.5-1.5B-Inst & Llama-3.1-8B-Inst & 44.2(17.6) & 34.1 & 38.4 & 76.1 & 22.8(16.9) & 26.5(16.5) & 24.7 & 16.5 & 22.2 \\
  \cline{1-1}
  Llama-3.1-8B-Inst & DeepSeek-V3 & \textbf{50.8}(23.2) & \textbf{39.8} & 37.8 & 37.3 & \textbf{27.8}(\textbf{21.7}) & \textbf{32.1}(\textbf{21.6}) & \textbf{31.1} & \textbf{27.4} & 13.2 \\
  Qwen2.5-7B-Inst & DeepSeek-V3 & 50.7(\textbf{23.4}) & 39.4 & \textbf{38.1} & \textbf{44.5} & 26.8(20.4) & 30.2(19.6) & 30.4 & 27.0 & \textbf{18.6} \\
  \bottomrule
  \end{tabular*}
  }
  \captionsetup{font=large}
  \caption{Comparison of different routers on ASQA, QAMPARI, and ELI5 datasets.} \label{tab:router_generalization}
\end{table*}

\subsection{Further Analysis}
\label{sec:b7}

\begin{table*}[h]
  \centering
  \begin{tabular}{@{}lccccc@{}}
  \toprule
  \textbf{DST Depth} & \textbf{Num} & \textbf{Standard RAG (EM)} & \textbf{RDR\textsuperscript{2} (EM)} & \textbf{$\Delta$EM} \\
  \midrule
  {[}1,2{]} & 62 & 52.6 & 53.0 & +0.4 \\
  (2,4] & 813 & 49.8 & 44.9 & +4.1 \\
  (4,7] & 73 & 32.8 & 43.6 & +10.8 \\
  \bottomrule
  \end{tabular}
  \captionsetup{font=large}
  \caption{Analysis of document depth and performance.} \label{tab:depth_analysis}
\end{table*}

We further provide details of our method from the perspective of time delay, computing budget and hierarchy modeling.

1) The offline DST construction is a deliberate design choice to achieve real-time retrieval efficiency, analogous to how standard RAG pipelines require offline processing steps like FAISS index building. In our experiments, constructing DSTs for the entire Wikipedia dump (~5.82M documents) takes approximately 20 minutes with parallelization across 8 CPU cores.

2) We evaluate the computing efficiency on ASQA dataset. Specifically, with single-turn routing, our method achieves +3.9\% EM over standard RAG while adding minimal computational cost (+0.779k / 0.017k input/output tokens). This low latency is primarily attributed to our RST design, concise router input/output format and parallelizable top-k retrieval/routing, before a single-turn final answer generation. Moreover, when employing multi-turn expansion, EM gains increase significantly to +10.8\% with modest overhead (+2.031k / 0.041k input/output tokens). Crucially, this performance-efficiency trade-off is tunable via the expand-iter hyperparameter, requiring no retraining.

3) For the hierarchy modeling, the Document Structure Tree (DST) can naturally handle both documents with rich structure and documents without a clear hierarchy, representing them as simple trees (e.g., single-level structures for flat documents). Our supplementary analysis shows that such cases are relatively rare in practice: on ASQA with Wikipedia documents, only 6.5\% of retrieved documents are shallow/flat (depth $\leq$ 2), where our method still yields a +0.4 EM improvement over standard RAG. As shown in Table\ref{tab:depth_analysis}, for the majority of documents (85.8\% with depth 2-4), we observe substantial improvements (+4.1 EM). For deeply structured documents (depth > 4), the gains are even more significant (+10.8 EM).

\clearpage

\clearpage

\section{Prompts}
\label{sec:c}

We show the detailed prompts for data curation, routing and inference as follows:

\begin{twocolprompt}{Prompt C.1: Training data curation prompt}
You are an expert in reading comprehension tasked with identifying relevant paragraphs from a document tree to answer a question. Follow these steps carefully:\\

1. Strict Relevance Assessment:

* First determine if the document's root heading is fundamentally relevant to the question.

* If the document is clearly about a different topic, immediately return "Cannot answer".

* Only proceed if the document is relevant or potentially relevant to the question.\\

2. Comprehensive Answer Extraction:

* For expanded paragraphs (visible content):

\ \ - Tag as "answer" ONLY if the paragraph DIRECTLY and COMPLETELY answers the question.
  
\ \ - If multiple paragraphs together provide a complete answer, tag ALL relevant ones.
  
\ \ - When paragraphs contain conflicting or supplementary information, include all that are relevant.\\

3. Collapsed Heading Expansion:

* If any unexpanded nodes might contain information that can answer the question? Tag as "expand" when ANY of these are true:

\ \ - The heading contains synonyms or standard terminology related to the question.
  
\ \ - The section appears in the expected position within a standardized document structure.
  
\ \ - Expanded sibling sections under the same parent contain answers.\\

4. Output Requirements:

* Strictly use this JSON format:

\ \ [
  
\ \ \ \ \{
    
\ \ \ \ \ \ "id": [integer],
      
\ \ \ \ \ \ "tag": "answer"|"expand",
      
\ \ \ \ \ \ "explanation": "[concise rationale]"
      
\ \ \ \ \}
    
\ \ ]
  
\ \ OR "Cannot answer".
  
* Never include irrelevant paragraphs just because they mention similar keywords.

* For multi-part answers, include ALL relevant paragraphs.

* If no paragraphs meet the strict criteria, return "Cannot answer".\\

\#\# Question\\
\{question\}\\

\#\# Document\\
\{context\}
\end{twocolprompt}

\begin{twocolprompt}{Prompt C.2: Routing module prompt}
You are asked to identify relevant nodes in a document tree that can answer the given question. Use [ANSWER] if a paragraph directly contributes to answering the question. Use [EXPAND] if a collapsed heading might contain information that can answer the question. If neither exists, reply exactly "Cannot answer".\\
\#\# Question\\
\{question\}\\

\#\# Document\\
\{context\}\\

\#\# Response
\end{twocolprompt}

\clearpage

\begin{twocolprompt}{Prompt C.3: Reader prompt for TriviaQA\, HotpotQA}
Instruction: Provide one accurate answer for the given question. Do not explain yourself or output anything else. \\
\#\# Paragraph\\
\{paragraph\}\\

\#\# Question\\
\{question\}\\

\#\# Response
\end{twocolprompt}

\begin{twocolprompt}{Prompt C.4: Reader prompt for QAMPARI}
Instruction: Provide a list of accurate answers for the given question. Separate answers by commas. Do not explain yourself or output anything else. \\
\#\# Paragraph\\
\{paragraph\}\\

\#\# Question\\
\{question\}\\

\#\# Response
\end{twocolprompt}

\begin{twocolprompt}{Prompt C.5: Reader prompt for ASQA\, ELI5}
Instruction: Write an accurate, engaging, and concise answer for the given question. Use an unbiased and journalistic tone.\\
\#\# Paragraph\\
\{paragraph\}\\

\#\# Question\\
\{question\}\\

\#\# Response
\end{twocolprompt}


\section{Demonstrations of RDR\textsuperscript{2}}
\label{sec:d}

We show a complete demonstration of our \textbf{RDR\textsuperscript{2}} as follows, including comparison of generation and retrieval stage and detailed routing actions.

\begin{table*}[!]
  \centering
  \resizebox{\textwidth}{!}{  
  \begin{tabularx}{0.9\textwidth}{@{\extracolsep{\fill}}lX}
  \toprule
    \textbf{User Question} & What is the tallest ride at six flags over texas? \\
    \textbf{Short Answers} & \textit{The Titan}, \textit{The New Texas Giant}, \textit{The Texas Giant}, \textit{Texas SkyScreamer}, \textit{Superman: Tower of Power} \\
    \textbf{Long Answer} & The Six Flags Over Texas theme park has a number of different rides and roller coasters. The park's tallest swing ride is the \textit{Texas SkyScreamer} at 400 feet and tallest triple drop tower ride is \textit{Superman: Tower of Power} at 325 feet. The tallest roller coaster at the park in 2001 was \textit{The Titan} at 245 feet. The park's tallest wooden roller coaster in 2011 was \textit{The New Texas Giant} at 153 feet, while its tallest one in 1990 was \textit{The Texas Giant} at 143 feet. \\
  \midrule
    \textbf{No-Retrieval Output} & At Six Flags Over Texas, thrill-seekers can experience the park's tallest ride, the '\colorbox{magenta}{Mr. Freeze: Reverse Blast}'. However, more recently, it has been surpassed by the '\colorbox{magenta}{Dare Devil Dive}' in the 'Thrill Zone' area of the park, but that is no longer there. \\
  \midrule
    \textbf{RAG Output} & According to the information provided, Six Flags Over Texas has had several record-breaking attractions over the years. In 2003, the park opened the \colorbox{yellow}{\textit{Superman Tower of Power}}, which was the tallest ride of its kind in the world at the time of its opening. However, it does not specify if it is still the tallest ride in the park. \\
  \midrule
    \textbf{$\text{RDR}^2$ Output} & According to the information provided, the tallest ride at Six Flags Over Texas is the \colorbox{yellow}{\textit{Texas SkyScreamer}}, which stands at a height of 400 feet. In 2003, the park opened the \colorbox{yellow}{\textit{Superman Tower of Power}}, which was the tallest ride of its kind in the world at the time. It's also worth noting that the park has had several record-breaking roller coasters, including \colorbox{yellow}{\textit{the Titan}}, which is the tallest, fastest, and longest coaster in Texas.\\
  \bottomrule
  \end{tabularx}
  }
  \captionsetup{font=large}
  \caption{End-to-end comparison between three frameworks.}
\end{table*}

\begin{table*}
  \centering
  \resizebox{\textwidth}{!}{  
  \begin{tabularx}{0.9\textwidth}{@{\extracolsep{\fill}}lX}
  \toprule
    \textbf{Top-3 Retrieved Chunks} & \\
  \midrule
    \textbf{Six Flags Over Texas} & \textcolor{red}{bringing back the past when they reopened Casa Magnetica.} In 2003 Six Flags Over Texas opened the \colorbox{yellow}{\textit{Superman Tower of Power}}. This was the tallest ride of its kind in the world at the time of its opening. [...] \textcolor{red}{For 2007 and 2008, Six Flags Over Texas was home to "Cirque',} \\
  \cline{1-1}
    \textbf{Six Flags Over Texas} & \textcolor{red}{end of the decade, Six Flags Over Texas had added ten roller coasters to its list of attractions.} During the first decade of the 21st century, Looney Tunes USA was restructured. In 2001, the park introduced its tallest, fastest, longest roller coaster, \colorbox{yellow}{\textit{Titan}}. [...] \textcolor{red}{The park has also made steps toward} \\
  \cline{1-1}
    \textbf{Titan (roller coaster)} & \colorbox{yellow}{\textit{Titan}} is a steel hyper coaster located at Six Flags Over Texas in Arlington, Texas. Unlike most hypercoasters, Titan is a combination of an out and back roller coaster and a twister roller coaster. It stands at 245 feet and contains a 255 drop at 85 miles per hour. It is the tallest, fastest, and longest coaster in Texas. [...] \textcolor{red}{In August 2000, Six} \\
  \midrule
    \textbf{Corresponding Routed Chunks} & \\
  \midrule
    \textbf{Six Flags Over Texas} & During the first decade of the 21st century, Looney Tunes USA was restructured. In 2001, the park introduced its tallest, fastest, longest roller coaster, Titan. [...] In 2003 Six Flags Over Texas opened the \colorbox{yellow}{\textit{Superman Tower of Power}}. This was the tallest ride of its kind in the world at the time of its opening. [...] \\ 
     & \hspace{1em} \textcolor{blue}{* Tallest Roller Coaster in Texas - \colorbox{yellow}{\textit{Titan}} (245ft)} \\
     & \hspace{1em} \textcolor{blue}{* Tallest swing ride in the world \colorbox{yellow}{\textit{Texas Skyscreamer}} (400ft) (2013)} \\
  \cline{1-1}
    \textbf{Titan (roller coaster)} & \colorbox{yellow}{\textit{Titan}} is a steel hyper coaster located at Six Flags Over Texas in Arlington, Texas. Unlike most hypercoasters, Titan is a combination of an out and back roller coaster and a twister roller coaster. It stands at 245 feet and contains a 255 drop at 85 miles per hour. It is the tallest, fastest, and longest coaster in Texas. \\
  \bottomrule
  \end{tabularx}
  }
  \captionsetup{font=large}
  \caption{Comparison between retrieved and routed chunks.}
\end{table*}

\begin{table*}
  \centering
  \renewcommand{\arraystretch}{0.9}
  \resizebox{\textwidth}{!}{  
  \begin{tabularx}{0.9\textwidth}{@{\extracolsep{\fill}}lX}
  \toprule
    & \textbf{Document Structure Tree} \\
  \midrule
    & -1: Six Flags Over Texas \\
    & \hspace{1em} 0: \textcolor{gray}{==Introduction==} \\
    & \hspace{2em} 1: \textcolor{gray}{Six Flags Over Texas is a 212-acre (86 ha) theme park located in Arlington, Texas, east of Fort Worth and about 15 miles (24km) west of Dallas. [...]} \\
    & \hspace{2em} 2: \textcolor{gray}{The park is managed by the Six Flags Entertainment Corp., which also owns 53.1\% interest of the Texas Limited Partnership that owns the park. [...]} \\
    & \hspace{1em} 3: ==History== \\
    & \hspace{2em} 4: \textcolor{gray}{===Initial planning and construction===} \\
    & \hspace{2em} \textcolor{gray}{[...]} \\
    & \hspace{2em} 16: \textcolor{gray}{===1990s===} \\
    & \hspace{3em} 17: \textcolor{gray}{The 1990s was a rather rough decade in comparison from decades past. The decade started off with a bang when Six Flags Over Texas introduced the Texas Giant roller coaster. [...]} \\
    & \hspace{2em} 18: ===2000s=== \\
    & \hspace{3em} 19: During the first decade of the 21st century, Looney Tunes USA was restructured. In 2001, the park introduced its tallest, fastest, longest roller coaster, \colorbox{yellow}{\textit{Titan}}. [...] In 2003 Six Flags Over Texas opened the \colorbox{yellow}{\textit{Superman Tower of Power}}. This was the tallest ride of its kind in the world at the time of its opening. [...] \\
    & \hspace{2em} 20: \textcolor{gray}{===2010s===} \\
    & \hspace{1em} 29: ==Firsts, bests, and other records== \\
    & \hspace{2em} 30: \textcolor{gray}{===Firsts and ones of a kind===} \\
    & \hspace{2em} 40: ===Records=== \\
    & \hspace{3em} 41: * Tallest Roller Coaster in Texas - \colorbox{yellow}{\textit{Titan}} (245ft) \\
    & \hspace{3em} 42: \textcolor{gray}{* Fastest Roller Coaster in Texas - \colorbox{pink}{Titan} (85mph)} \\
    & \hspace{3em} 43: \textcolor{gray}{* Largest Land Based Oil Derrick - \colorbox{pink}{Oil Derrick} (300ft)} \\
    & \hspace{3em} 44: * Tallest swing ride in the world \colorbox{yellow}{\textit{Texas Skyscreamer}} (400ft) (2013) \\
    & \hspace{2em} 45: \textcolor{gray}{===Awards===} \\
    & \hspace{1em} 48: \textcolor{gray}{==Events==} \\
    & \hspace{1em} 54: \textcolor{gray}{==Areas and attractions==} \\
    & \hspace{2em} 56: \textcolor{gray}{===Star Mall===} \\
    & \hspace{2em} \textcolor{gray}{[...]} \\
    & \hspace{2em} 157: \textcolor{gray}{===Tower===} \\
    & \hspace{1em} 168: \textcolor{gray}{==Former Attractions==} \\
  \midrule
    & \textbf{Routing Actions} \\
  \midrule
      & Light content node \textit{17} from retrieved passages. \\
    1 & \textbf{[EXPAND]} \textit{0} \\
      & Light content node \textit{1}, \textit{2} from expand action. \\
    2 & \textbf{[REFUSE]} \\
      & Light content node \textit{19} from retrieved passages. \\
    3 & \textbf{[ANSWER]} \textit{19} \textbf{[EXPAND]} \textit{40} \\
      & Light content node \textit{41}, \textit{42}, \textit{43}, \textit{44} from expand action. \\
    4 & \textbf{[ANSWER]} \textit{41}, \textit{44} \\
  \midrule
    &  \textbf{Routing Passages: }  \textit{19},  \textit{41},  \textit{44} \\
  \bottomrule
  \end{tabularx}
  }
  \captionsetup{font=large}
  \caption{Demonstration of routing actions.}
\end{table*}

\end{document}